\documentclass[dvipsnames]{article}

\usepackage{iclr2027_conference,times}
\usepackage{amsmath,amssymb,amsthm}
\usepackage{fix-cm} %
\usepackage{booktabs, xcolor}
\usepackage{float}
\usepackage{graphicx}
\usepackage[hidelinks,backref=page,colorlinks=True,linkcolor=BrickRed,citecolor=Blue]{hyperref}
\usepackage{url}
\usepackage{cleveref, mathtools}
\usepackage{tikz}
\usetikzlibrary{arrows.meta,positioning}
\usepackage{subcaption}
\usepackage{wrapfig}
\usepackage{enumitem}
\usepackage{pifont}
\floatstyle{ruled}
\newfloat{algorithm}{tbp}{loa}
\floatname{algorithm}{Algorithm}
\crefname{algorithm}{algorithm}{algorithms}
\crefname{section}{\S}{\S}
\crefname{subsection}{\S}{\S}
\Crefname{algorithm}{Algorithm}{Algorithms}

\newtheorem{assumption}{Assumption}
\newtheorem{theorem}{Theorem}
\newtheorem{proposition}{Proposition}
\newtheorem{lemma}{Lemma}

\crefname{assumption}{Assumption}{Assumptions}
\Crefname{assumption}{Assumption}{Assumptions}
\crefname{equation}{Eq.}{Eqs.}
\Crefname{equation}{Eq.}{Eqs.}
\crefname{figure}{Fig.}{Figs.}
\Crefname{figure}{Fig.}{Figs.}
\theoremstyle{definition}

\theoremstyle{plain}

\definecolor{muone}{HTML}{009E73}
\definecolor{mutwo}{HTML}{D55E00}
\definecolor{muthree}{HTML}{CC79A7}
\definecolor{causalblue}{HTML}{0072B2}
\definecolor{contrasttwo}{HTML}{7E3F98}
\definecolor{contrastthree}{HTML}{677E96}
\definecolor{nuisance}{HTML}{17202E}

\newcommand{\E}{\mathbb{E}}
\newcommand{\Var}{\mathop{\mathrm{Var}}}
\newcommand{\Cov}{\mathop{\mathrm{Cov}}}
\newcommand{\indep}{\mathrel{\perp\!\!\!\perp}}
\newcommand{\1}{\mathbf{1}}
\newcommand{\R}{\mathbb{R}}

\newcommand{\methodname}{Outstrumental Variable Regression}
\newcommand{\methodabbr}{OVR}
\newcommand{\restrictionname}{inclusion restriction}
\newcommand{\restrictionnamecap}{Inclusion restriction}
\newcommand{\restrictionabbr}{IR}

\definecolor{cbblue}{HTML}{0072B2}
\definecolor{cborange}{HTML}{E69F00}
\definecolor{cbgreen}{HTML}{009E73}

\title{\methodname
}
\author{Felix Schur$^1$\footnotemark[2] \quad Linus K\"uhne$^{1,2}$ \quad
Julius von K\"ugelgen$^1$\thanks{Joint supervision.} \quad
Jonas Peters$^1$\footnotemark[1]\\
\normalfont $^1$Seminar for Statistics, ETH Zurich\\
\normalfont $^2$ETH AI Center, ETH Zurich}

\iclrfinalcopy

\begin{document}
\maketitle
\begingroup
\renewcommand{\thefootnote}{\fnsymbol{footnote}}
\footnotetext[2]{Corresponding author: \texttt{felix.schur@stat.math.ethz.ch}.}
\endgroup
\fancyhead{}

\begin{abstract}
\looseness-1 
Estimating the causal effect of a treatment $X$ on an outcome $Y$ from observational data in the presence of unobserved confounding requires additional structure or assumptions. 
Existing identification strategies
such as instrumental variable regression 
exploit exogenous variation in only the treatment. We show that the causal effect can also be identified if the outcome mechanism varies across a finite set of environments, even if the treatment distribution is unchanged. Concretely, we introduce \emph{\methodname} (\textsc{\methodabbr}), which
proceeds in three steps: (i)~compute the conditional expectation of $Y$ given $X$ in each environment; (ii)~form the vector of differences in these conditional expectations between all environments and a reference environment
(we call this vector \textit{contrast});
and (iii)~regress $Y$ on $X$ while adjusting for the contrast and 
environment. We prove that this yields the true causal effect if the following two conditions hold: \emph{\MakeLowercase{\restrictionname}}, i.e., the contrast determines the conditional confounding bias; and \emph{positivity}, i.e., the treatment varies sufficiently strongly after adjustment. We prove that a finite-sample version of \textsc{\methodabbr} is consistent under regularity conditions and demonstrate the procedure's effectiveness on synthetic and real-world data.
\end{abstract}

\section{Introduction}
\label{sec:intro}
\looseness-1 
Estimating the causal effect of a treatment $X$ on an outcome
$Y$ is a central goal of empirical science. Randomized controlled trials~\citep{fisher1935design}
provide a direct solution, but are often infeasible, unethical, or too costly. 
When only observational data is available, unobserved confounding 
usually renders
causal effects unidentifiable without additional assumptions.

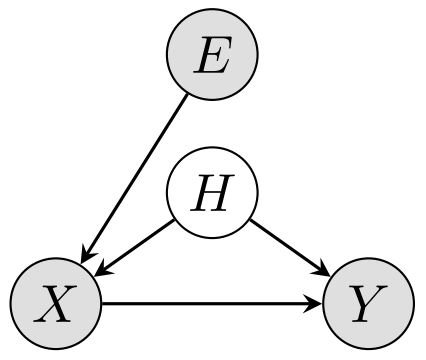
\begin{wrapfigure}{r}{0.21\textwidth}
\vspace{-1.5em}
\centering
\captionsetup{font=footnotesize,skip=3pt,justification=centering}
\begin{tikzpicture}[
    node distance=3mm and 6mm,
    observed/.style={draw, circle, minimum size=6mm, inner sep=.25pt, , fill=gray!25},
    hidden/.style={draw, %
    circle, minimum size=6mm, inner sep=.25pt},
    arrow/.style={-{Latex[length=2mm]}, semithick}
]
\node[hidden] (hid) {$H$};
\node[observed, above=of hid] (env) {$E$};
\node[observed, below left=of hid] (exp) {$X$};
\node[observed, below right=of hid] (out) {$Y$};
\draw[arrow, -stealth] (env) -- (exp);
\draw[arrow, -stealth] (hid) -- (exp);
\draw[arrow, -stealth] (hid) -- (out);
\draw[arrow, -stealth] (exp) -- (out);
\end{tikzpicture}
\caption{IV setting.}
\label{fig:iv-setting}
\vspace{-1em}
\end{wrapfigure}
One of the most
widely used approaches to this problem is instrumental variable (IV) regression, which
has played a central role in estimating causal effects from natural
experiments and policy changes~\citep{angrist1996identification,angrist2009mostly}.
In graphical terms, an instrument~$E$ supplies a 
`treatment channel':
it generates variation in $X$
(via $E\to X$ in~\Cref{fig:iv-setting}), 
while 
the
exclusion restriction rules out
alternative pathways from $E$ to $Y$, captured in~\cref{fig:iv-setting} as $E\not\to H$ and $E\not\to Y$~\citep{pearl2009causality, peters2017elements}.

\begin{figure}[t]
\centering
\includegraphics[width=\linewidth]{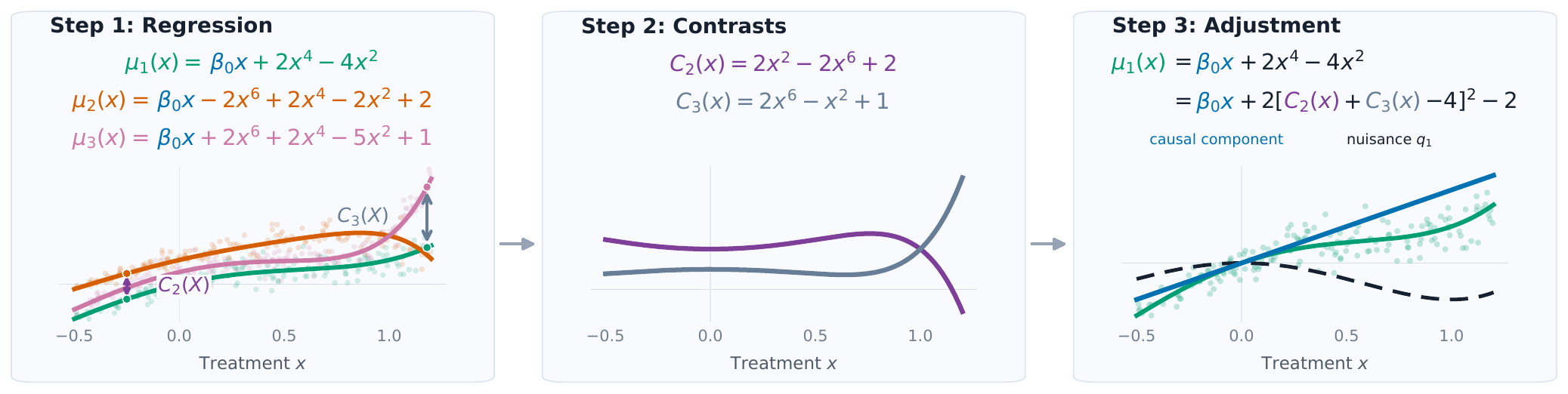}
\caption{\looseness-1
\methodname{} (\methodabbr{}). We illustrate the main steps  of OVR using three environments.
\textbf{Step 1: Regression.} Estimate the conditional expectations $\textcolor{muone}{\mu_1}, \textcolor{mutwo}{\mu_2}, \textcolor{muthree}{\mu_3}$ of $Y$ given $X$ in each environment.
\textbf{Step 2: Contrast.} Subtract $\textcolor{muone}{\mu_1}$ from $\textcolor{mutwo}{\mu_2}$ and $\textcolor{muthree}{\mu_3}$ to obtain the contrasts~$\textcolor{contrasttwo}{C_2}$ and~$\textcolor{contrastthree}{C_3}$; the shared causal contribution $\textcolor{causalblue}{\beta_0x}$ cancels.
\textbf{Step 3: Adjustment.} 
Regress~$Y$ on $X$, adjusting 
for the contrast vector $C(X) \coloneqq (\textcolor{contrasttwo}{C_2(X)}, \textcolor{contrastthree}{C_3(X)})$ and environment $E$. This recovers~
$\textcolor{causalblue}{\beta_0}$ if
$C(X)$ determines the nuisance $q_1(x)\coloneqq \textcolor{muone}{\mu_1(x)}-\textcolor{causalblue}{\beta_0x}$ (dashed black) and enough variation in~$X$ remains after 
partialling out $C(X)$ and $E$.
}
\label{fig:ovr-illustration}
\end{figure}

\WFclear
\begin{wrapfigure}{r}{0.21\textwidth}
\vspace{-.1em}
\centering
\captionsetup{font=footnotesize,skip=3pt,justification=centering}
\begin{tikzpicture}[
    node distance=3mm and 6mm,
    observed/.style={draw, circle, minimum size=6mm, inner sep=.25pt, , fill=gray!25},
    hidden/.style={draw, %
    circle, minimum size=6mm, inner sep=.25pt},
    arrow/.style={-{Latex[length=2mm]}, semithick}
]
\node[hidden] (hid) {$H$};
\node[observed, above=of hid] (env) {$E$};
\node[observed, below left=of hid] (exp) {$X$};
\node[observed, below right=of hid] (out) {$Y$};
\draw[arrow, -stealth] (env) -- (out);
\draw[arrow, dashed, -stealth] (env) -- (hid);
\draw[arrow, dashed, -stealth] (env) -- (exp);
\draw[arrow, -stealth] (hid) -- (exp);
\draw[arrow, -stealth] (hid) -- (out);
\draw[arrow, -stealth] (exp) -- (out);
\end{tikzpicture}
\caption{\methodabbr{} setting.}
\label{fig:ovr-setting}
\vspace{-.15em}
\end{wrapfigure}
\looseness-1 We propose a complementary identification approach that exploits
a different source of information: variation in the
`outcome channel',
that is, $E\to Y$ (see \Cref{fig:ovr-setting}).
Assume $E$ is categorical and corresponds to different environments in which we observe the data (for example, different
sites, time periods, experimental conditions, or policy regimes).
The intuition behind our method is as follows: because the causal effect is invariant, changes in the $X$--$Y$ relationship across environments (the contrasts) can reveal what the confounding contribution looks like (i.e., the `shape' of the confounding). 
As we will see, we can therefore remove the confounding by adjusting for the contrast and the environment. More precisely,
our method proceeds in three
steps (see~\cref{fig:ovr-illustration} for an overview). First, we estimate $\mu_e(x)\mathrel{\coloneqq}\E[Y| X=x,E=e].$
Second, we compute a vector of differences between these conditional expectations relative to
an arbitrary reference environment~$e_0$,
$
    C(x)
    \coloneqq 
    (\mu_e(x)-\mu_{e_0}(x))_{e\neq e_0}.
$
Under the assumption that the causal effect is invariant across environments, its contribution
cancels in these differences. What remains therefore carries information
about the confounding that biases the observational relationship between
$X$ and~$Y$. Third, we use $C(X)$, together with the
environment $E$, as an adjustment variable. We call this
procedure \emph{\methodname} (\textsc{\methodabbr}).

Similar
to IV, our identifying assumptions, 
\Cref{ass:cs} and 
\Cref{ass:pos},
are not primarily graphical restrictions
on which causal paths may or may not be present. Instead, they constrain how the outcome
mechanism varies across environments. This allows us to leave much of the
remaining data-generating process unrestricted, without specifying structural
equations for all variables.
Our identifying assumptions are thus not implied by the graph in~\Cref{fig:ovr-setting} alone; in
\Cref{sec:summary}, we give three concrete structural settings in which they
hold. 

\looseness-1 \textsc{\methodabbr} is complementary to multi-environment methods
that exploit environment-induced variation in the treatment
\citep{rothenhausler2019causal,rothenhausler2021anchor,
schur2026many,gnecco2023boosted}.
Thus, \textsc{\methodabbr} can identify the causal coefficient even when the treatment distribution is unchanged across environments. Conversely, IV can identify effects in settings where the outcome contrasts are uninformative. Neither set of identifying conditions contains the other.

\textbf{Contributions.} We establish three results. First, under the \restrictionname~(\Cref{ass:cs}) and positivity (\Cref{ass:pos}) the causal effect is identified by  \textsc{\methodabbr} (\Cref{thm:id}). Second, we highlight three structural settings which imply both of these assumptions, and provide a sensitivity bound that quantifies the bias when the \restrictionname{} does not hold~%
(\Cref{sec:summary,sec:sensitivity}). Third, adding Gaussian noise to smooth the empirical version of the contrasts estimator yields a consistent estimator
(\Cref{thm:main-consistency}), which we evaluate %
on six synthetic settings and a real-data benchmark~(\Cref{sec:experiments}).%

\section{Causal reasoning %
via~\methodname}
\label{sec:tracer}
We first formalize the  problem setting~(\cref{sec:problem_setting}) and build intuition through an illustrative example~(\cref{sec:warm_up}). We then describe the proposed approach for the general case~(\cref{sec:general_algorithm}). %
\subsection{Problem setting}
\label{sec:problem_setting}
\looseness-1 \textbf{Setting.} We observe realizations of the random variables $(X,Y,E)\sim P$, where $X\in\R^d$ is a (possibly multivariate) treatment, $Y\in \R$ is an outcome,
and $E\in\mathcal{E}$ is an %
environment label. Throughout, we assume %
that $\mathcal{E}$ is a finite set. 
We assume
that the average causal effect of $X$ on~$Y$ is linear and
invariant across environments; more precisely,
there exists  some $\beta_0\in\R^d$ and environment-specific
intercepts $(\alpha_e)_{e\in\mathcal E}$ such that for every $e\in\mathcal E$ and $P_X$-almost all
$x$,
\begin{equation}
\label{eq:invariant_causal_effect}
\E[Y\mid\operatorname{do}(X=x),E=e]
=
\beta_0^\top x + \alpha_e.
\end{equation}
Furthermore, we assume environment overlap:
 for $P_X$-almost every $x$ and every $e\in\mathcal{E}$,
\begin{equation}
\label{eq:environment_overlap}
P(E=e\mid X=x)>0.    
\end{equation}
We also assume $\E\|X\|_2^2+\E[Y^2]<\infty$ to ensure that second moments are finite.

\textbf{Goal.} %
Our goal is to identify $\beta_0$ from the (observational) joint distribution $P$ and estimate it from an i.i.d.\ sample of~$P$. We do not assume access to interventional data. %
With unobserved confounders~$H$, the expression in~\cref{eq:invariant_causal_effect} generally differs from the observational conditional expectation ${\E[Y|X=x,E=e]}$. Simply regressing  $Y$ on $X$
(after stratifying
or
pooling environments) %
thus generally yields 
an inconsistent
estimator of~$\beta_0$.

\looseness-1 \textbf{Graphical perspective.} The considered problem setting includes, for example, data-generating processes captured by a structural causal model~\citep[SCM;][]{pearl2009causality, bongersFoundationsStructuralCausal2021} whose induced graph is shown in \Cref{fig:ovr-setting} (where we allow for the dashed arrows) %
and with structural equation for $Y$ given by
\begin{equation}
\label{eq:scm_general}
    Y
    \coloneqq 
    \beta_0^\top X
    + \alpha_E
    + \gamma(H,E,X,U_Y),
\end{equation}
\looseness-1 where $U_Y$ is unobserved noise and, 
for every fixed $x$,
$\E[\gamma(H,E,x,U_Y)|E]=0$ almost surely.
 Thus, our setting %
requires neither the treatment distribution 
nor the %
confounder distribution to be invariant across environments.

\subsection{%
An illustrative example}
\label{sec:warm_up}
We %
first illustrate the main idea behind 
our 
identification strategy 
through a simple example with two environments
$\mathcal{E}=\{1,2\}$ satisfying $P(E=1),P(E=2)>0$. 
Assume $X$, $Y$, and $H$
are real-valued and
centered 
and consider %
the structural model
\begin{equation}
\label{eq:scm_example}
    Y\coloneqq \beta_0%
    X+\lambda_E H+U_Y,
\end{equation}
where $U_Y$ satisfies
$\E[U_Y|X,E]=0$.
\looseness-1 We assume $\lambda_1 \neq \lambda_2$,
so the environment influences the effect of $H$ on $Y$ (via $\lambda_E$).
But we assume the environment does not change the joint distribution of $(H,X)$: that is,
we have $E\indep(H,X)$, corresponding to the solid-arrow subgraph of \Cref{fig:ovr-setting}. This is a special case of \eqref{eq:scm_general} with
$\alpha_e=0$ and $\gamma(H,e,x,U_Y)=\lambda_eH+U_Y$,
and hence satisfies~\eqref{eq:invariant_causal_effect}
without %
environment-specific intercepts.
In population form, our approach consists of %
three steps.

\textbf{Step 1: Regression.} Compute the conditional expectation of $Y$ given $X=x$ in environment $e$,
\begin{equation}
\label{eq:mu_e_example}
    \mu_e(x)\coloneqq\E[Y\mid X=x,E=e]\overset{\eqref{eq:scm_example}}{=}\beta_0 x+\lambda_e\E[H\mid X=x],
\end{equation}
where we have used $\E[U_Y|X,E]=0$ and $H\indep E \mid X$.

\textbf{Step 2: Contrast.} 
Form the contrast in the means from~\eqref{eq:mu_e_example} across environments, 
\begin{equation}
\label{eq:contrast_example}
    C(x)\coloneqq \mu_2(x)-\mu_1(x) = (\lambda_2-\lambda_1)\E[H\mid X=x].
\end{equation}
The invariant causal effect $\beta_0 x$ cancels from the contrast $C(x)$, while the conditional confounding signal $\E[H|X=x]$ that biases the observational conditional expectation is retained.

\textbf{Step 3: Adjustment.}
Use $C(X)$ and $E$ for adjustment. 
To do so, 
define the residualized variables
\begin{equation}
\label{eq:residualized_variables}
    \widetilde X
    \coloneqq
    X-\E[X\mid C(X),E] 
    \qquad 
    \text{and}
    \qquad 
    \widetilde Y
    \coloneqq
    Y-\E[Y\mid C(X),E].
\end{equation}
Taking conditional expectations of~\eqref{eq:scm_example} given $(C(X),E)$ yields
\begin{equation}
\label{eq:cond_exp_example}
    \E[Y\mid C(X),E] = \beta_0%
    \E[X\mid C(X),E] + \lambda_E\E[H\mid X],
\end{equation}
where we used that $E\!\indep\!(H,X)$ and $\E[H|C(X)]\!=\!\E[H|X]$ by~\eqref{eq:contrast_example}, as well as ${\E[U_Y|C(X),E]\!=\!0}$ due to the tower property and $\E[U_Y|X,E]\!=\!0$.
Subtracting~\eqref{eq:cond_exp_example} %
from~
\eqref{eq:scm_example} %
then yields%
\begin{equation*}
    \widetilde Y
    =
    \beta_0%
    \widetilde X
    +
    \lambda_E\big(H-\E[H\mid X]\big)
    +
    U_Y.
\end{equation*}
Partialling out $C(X)$ and $E$ thus removes exactly the part of the confounding that is predictable from~$X$.
The last two terms have a conditional expectation of zero given $(X,E)$. Since $\widetilde X$ is a function of $(X,E)$, their product with $\widetilde X$ also has an expectation of zero by the tower property. Multiplying by $\widetilde X$ and taking expectations yields %
${
    \E[\widetilde X\widetilde Y]
    =
    \beta_0\E[\widetilde X^2]}.
$
Consequently, whenever $\E[\widetilde X^2]>0$, the causal parameter $\beta_0$ is identified by $\E[\widetilde X\widetilde Y]/\E[\widetilde X^2]$.
The contrast $C(X)$ can therefore be used to identify~$\beta_0$ as long as sufficient variation in $X$ remains after partialling out $C(X)$ and $E$.

\subsection{\methodname~(\methodabbr{})}
\label{sec:general_algorithm}
We now define the 
general
population-level procedure for the setting from~\Cref{sec:problem_setting}.
For each environment ${e\in\mathcal{E}}$, define 
$
    {\mu_e(x)\coloneqq \E[Y|X=x,E=e]}
$ as in~\eqref{eq:mu_e_example}
and the corresponding \emph{nuisance}
$
    {q_e(x)\coloneqq \mu_e(x)-\beta_0^\top x}
$.
Thus,
$
    \mu_e(x)=\beta_0^\top x+q_e(x),
$
where $q_e$ contains the environment-specific intercept~$\alpha_e$ from~\eqref{eq:invariant_causal_effect} and the difference between observational and interventional conditional expectations. 
(This decomposition does not
require specifying an underlying SCM for 
how
$(E,X)$ and potential hidden confounders are generated.)
Fix an arbitrary reference environment $e_0\in\mathcal{E}$ and define the \emph{contrast}
$C$ by
\begin{equation}
\label{eq:contrast}
    C(x)
    \coloneqq 
    \bigl(\mu_e(x)-\mu_{e_0}(x)\bigr)_{e\ne e_0}
    =
    \bigl(q_e(x)-q_{e_0}(x)\bigr)_{e\ne e_0}.
\end{equation}
The environment overlap condition~\eqref{eq:environment_overlap} ensures that every $\mu_e$, and hence $C$, is identified $P_X$-almost surely.
The invariant causal term $\beta_0^\top x$ cancels from each component of
$C(x)$.   \emph{\methodname{}}~(\methodabbr{})
adjusts for $C(X)$ and $E$ by residualizing with
respect to $(C(X),E)$.
With $\widetilde{X}$ and~$\widetilde{Y}$ defined as the residualized variables from~\cref{eq:residualized_variables},
let
\begin{equation}
\label{eq:population-tracer}
    \Sigma_C\coloneqq \E\left[\widetilde X\widetilde X^\top\right]
    \qquad \text{and} \qquad
    \beta_{\mathrm{\methodabbr}}\coloneqq \Sigma_C^{-1}\E\left[\widetilde X\widetilde Y\right],
\end{equation}
whenever $\Sigma_C$ is nonsingular (\Cref{ass:pos}).
For finite samples, to reduce sensitivity to contrast estimation errors, we introduce Smoothed \methodname{}
(\Cref{alg:smoothed-ccr}) and prove its consistency in~\Cref{sec:theory}.

\section%
{Identifiability and a consistent algorithm}
\label{sec:identifiability}
\looseness-1
We first give sufficient conditions for identifiability under which $\beta_{\mathrm{\methodabbr}}$ equals $\beta_0$ in~\cref{subsec:identifiability} and
then establish consistency of the proposed Smoothed \methodabbr{} estimator in~\cref{sec:theory}.
\subsection{Identifiability}
\label{subsec:identifiability}
Our identification strategy for the invariant causal effect from~\eqref{eq:invariant_causal_effect} rests on two key assumptions. First, we
require the baseline nuisance $q_{e_0}(X)$ to depend on $X$
only through the contrast $C(X)$. Since the contrasts give the
differences $q_e(X)\!-\!q_{e_0}(X)$, the full nuisance $q_E(X)$ is
then determined by $(C(X),E)$. Residualizing $X$ and $Y$ with
respect to $(C(X),E)$ therefore eliminates the nuisance term
from the outcome equation.
This is what happens in \Cref{sec:warm_up}: there,
$
    C(X) \!=\! (\lambda_2\!-\!\lambda_1)\E[H|X],
$
 so $\lambda_2\!\neq\!\lambda_1$ makes the contrast determine $\E[H|X]$ and hence the baseline nuisance ${q_1(X)\!=\!\lambda_1\E[H|X]}$.
\begin{assumption}[\restrictionnamecap{} (\restrictionabbr)]
\label{ass:cs}
There exists a measurable function $\varphi$ 
such that the baseline nuisance satisfies $q_{e_0}(X)=\varphi(C(X))$ almost surely.
\end{assumption}
\looseness-1
If the nuisance $q_e$ does not vary across environments, \restrictionabbr{} 
fails unless the nuisance is almost surely constant.
Validity of
\restrictionabbr{}
does not depend on the choice of reference environment
because, for all~$e$,
$q_e(X)=q_{e_0}(X)+C_e(X)$.
Furthermore, 
the contrasts corresponding to 
two different  choices of $e_0$
contain the same information,
so the population algorithm does not change, either.
\Cref{fig:ovr-illustration} illustrates \restrictionabbr{} with reference environment $e_0=1$: the two contrasts jointly determine the baseline nuisance $q_1$ through a nonlinear function (\Cref{fig:ovr-illustration}, right). Neither contrast alone determines $q_1$, and no affine function of both contrasts can represent it.

Second, we require
that enough variation in $X$ remains after conditioning on $(C(X),E)$.
\begin{assumption}[Positivity]
\label{ass:pos}
$\Sigma_C=\E[\widetilde X\widetilde X^\top]$ is positive definite.
\end{assumption}
Since, for all $v \in \mathbb R^d$, we have $v^\top\Sigma_Cv=\E[\Var(v^\top X|C(X),E)]$, positivity means that every nonzero linear combination of \(X\) has positive residual variance after conditioning on \((C(X),E)\).
Thus, the contrast is required to
determine the nuisance while leaving residual variation in every linear treatment direction. 
If $C$ is injective on the support of $X$, then $C(X)$ 
determines $X$: \restrictionabbr{} holds automatically, but $\smash[t]{\widetilde X=0}$ almost surely and $\Sigma_C=0$, thus violating positivity.
Positivity can be assessed from the observed data once $C$ has been estimated, since it concerns the residual covariance of $X$.
\restrictionabbr{} is different in this regard: it requires the contrasts to capture the unobserved confounding term and therefore cannot be verified from observational data alone.

\begin{theorem}[Identification]
\label{thm:id}
Under the setting of \Cref{sec:problem_setting}, \Cref{ass:cs,ass:pos} imply $\beta_{\mathrm{\methodabbr}}=\beta_0$.
\end{theorem}
In particular, $\beta_0$ is identifiable from the (observational) joint distribution $P$ of $(X,Y,E)$.
 Proofs of the main-text results are collected in \Cref{app:proofs}.

\textbf{Relaxing \restrictionabbr.
} \restrictionabbr, alongside positivity, is sufficient but not necessary for $\beta_{\mathrm{\methodabbr}}=\beta_0$. 
Define the nuisance remaining after adjustment by
$
r_E(X)\coloneqq q_E(X)-\E[q_E(X)|C(X),E].
$
Under positivity, $\beta_{\mathrm{\methodabbr}}=\beta_0$ if and only if $\E[\widetilde Xr_E(X)]=0$. \restrictionabbr{} implies the stronger condition $r_E(X)=0$ almost surely. 
More precisely,
in~\Cref{app:bias-without-ir}
we prove that 
$\beta_{\mathrm{\methodabbr}}-\beta_0=\Sigma_C^{-1}\E[\widetilde Xr_E(X)]$.

\textbf{Robustness analysis.}
\label{sec:sensitivity}
With a similar reasoning, 
we can bound the bias of OVR when \restrictionabbr{} fails to hold. 
Suppose
$q_{e_0}(X)\in L^2$ and define the strength of \restrictionabbr{} violation
\[
    \delta_{\mathrm{\restrictionabbr}}^2
    \coloneqq 
    \inf_\varphi
    \E\!\left[
        \bigl(q_{e_0}(X)-\varphi(C(X))\bigr)^2
    \right]
    =
    \E\!\left[
        \Var\!\left(q_{e_0}(X)\mid C(X)\right)
    \right],
\]
where the infimum is over measurable functions $\varphi$. 
Thus, $\delta_{\mathrm{\restrictionabbr}}$ measures the variation in the baseline nuisance
that is not explained by the contrasts, and \restrictionabbr{} holds if and only if
$\delta_{\mathrm{\restrictionabbr}}=0$.
Under positivity, 
we obtain the following
bound (see \Cref{app:bias-without-ir}):
\[
    \|\beta_{\mathrm{\methodabbr}}-\beta_0\|_2
    \le
    \frac{\delta_{\mathrm{\restrictionabbr}}}
         {\sqrt{\lambda_{\min}(\Sigma_C)}},
\]
where $\lambda_{\min}(\Sigma_C)$ is the smallest eigenvalue of $\Sigma_C$. Hence, the effect of violating \restrictionabbr{} depends on both its severity, measured by $\delta_{\mathrm{\restrictionabbr}}$, and positivity: small violations of \restrictionabbr{}
lead to small worst-case bias when $\lambda_{\min}(\Sigma_C)$ is bounded away from zero, but can be amplified when positivity is weak.

\textbf{Violations of positivity.
}
Positivity (\Cref{ass:pos}) fails when $C(X)$ determines a nonzero linear combination of $X$ (e.g., when $C$ is injective).
For example, if $\E[H|X]=AX$ and $C(X)$ recovers~$AX$, every direction in the row space of $A$ has zero residual variance.

\subsection[Consistency]{Smoothed OVR: A consistent algorithm}
\label{sec:theory}
Turning the population identification result into an estimator introduces an additional difficulty: the contrast $C(X)$ is itself unknown and must be estimated from the data. We use sample splitting to ensure independence between the estimated contrast and the sample used for the subsequent residualization step.
More importantly, consistency of the contrast  estimation alone 
may not be
sufficient because conditioning can be highly sensitive to small estimation errors.
For example, if 
$|\mathcal{E}| = 2$,
$C(X)=0$ and $\smash[t]{\widehat C(X)=\delta_n X}$ for scalar $X$, then 
for every $\delta_n\neq0$, $\smash[t]{\widehat C(X)}$ reveals~$X$ exactly 
and 
thus
`perfect' residualization yields
$X-\E[X\mid\widehat C(X),E]=0$,
so
OVR fails , even though $\smash[t]{\widehat C(X)\to C(X)}$ as $\delta_n\to0$.

\begin{algorithm}[t]
\caption[Smoothed \methodname]{Smoothed \methodname}
\label{alg:smoothed-ccr}
{
\textbf{Input:} Data $\{(X_i,Y_i,E_i)\}_{i=1}^n$, folds $K$, smoothing scale $\tau$, number of perturbations $M$.
\begin{enumerate}[leftmargin=*,nosep]
\item \textbf{Estimate contrasts.}
Split data into $\mathcal I_C$ and $\mathcal I_\beta$,
then split $\mathcal I_C$ into $\mathcal I_A$ and $\mathcal I_B$.
Fit contrasts $\widehat C^A$ and~$\widehat C^B$ on these
respective samples and set
$\overline C\coloneqq(\widehat C^A+\widehat C^B)/2$ and 
$D\coloneqq\widehat C^A-\widehat C^B$.

\item \textbf{Fit residualization regressions.}
Partition $\mathcal I_\beta$ into folds $I_1,\ldots,I_K$.
For each $k$, use ${I_{-k}\!\coloneqq\!\mathcal I_\beta\setminus I_k}$
to estimate
$\widehat\Omega_k\!\coloneqq\!
\widehat{\Cov}_{I_{-k}}(D(X))/4$
and fit regressions $\widehat f_{X,k}$ and $\widehat f_{Y,k}$
of $X$ and $Y$ on $(\overline C(X)\!+\!\xi,E)$, where we
draw $\xi\!\sim\! N(0,\tau^2\widehat\Omega_k)$ independently
for each training observation.

\item \textbf{Partial out on held-out folds.}
For each $k$, draw
$\xi'_{im}\sim N(0,\tau^2\widehat\Omega_k)$ independently
over $i\in I_k$ and $m=1,\ldots,M/2$.
Using the same perturbations for $X$ and $Y$, set
\[
\widehat Z_i\coloneqq Z_i-
\frac{1}{M}\sum_{m=1}^{M/2}\sum_{s\in\{-1,1\}}
\widehat f_{Z,k}\bigl(\overline C(X_i)+s\xi'_{im},E_i\bigr)
\qquad\text{for}\qquad Z\in\{X,Y\}.
\]
\end{enumerate}
\textbf{Return:}
$\widehat\beta_{\mathrm{SOVR}}\coloneqq
\left(\sum_{i\in\mathcal I_\beta}\widehat X_i\widehat X_i^\top\right)^{-1}
\sum_{i\in\mathcal I_\beta}\widehat X_i\widehat Y_i$.
}
\end{algorithm}

To limit 
unwanted
leakage of $X$ through the estimation error in $C(X)$, we 
add Gaussian noise to $\smash[t]{\widehat C(X)}$ at the scale of its estimation uncertainty. This makes the adjustment step less sensitive to information about $X$ that is present only due to estimation error. Less noise is added when the variability in contrast estimation is lower.
We refer to the resulting procedure, summarized in~\Cref{alg:smoothed-ccr}, as Smoothed \methodabbr{}.
Under consistent nuisance estimation and suitable calibration of the smoothing noise, \Cref{thm:main-consistency} establishes consistency of the resulting estimator. 
\Cref{app:theory} contains the 
precise
assumptions and the full theorem statement (\Cref{thm:consistency-full}); the proof is in \Cref{app:proof-consistency}.

\begin{theorem}[Consistency, informal]
\label{thm:main-consistency}
Under \restrictionabbr{} (\Cref{ass:cs}), positivity (\Cref{ass:pos}) and additional regularity conditions (\Cref{ass:est}),
Smoothed \methodabbr{}
(\Cref{alg:smoothed-ccr}) is consistent for $\beta_0$.
\end{theorem}

\section[Three special cases]{Three special cases}
\label{sec:summary}

We provide three special cases of the setting described in
\Cref{sec:problem_setting} for which the \restrictionname{} holds
and discuss the positivity condition (\Cref{ass:pos})
needed for identification in each.

\subsection[Multivariate latent with environment-specific loadings]{Multivariate latent with environment-specific loadings}
\label{sec:lowrank}
We now provide sufficient identifiability conditions for (a generalized version of) the motivating example from~\Cref{sec:warm_up}.
Let $X\in\R^d$, $H\in\R^s$, $E\in\{1,\ldots,J\}$, and $\lambda_1,\ldots,\lambda_J\in\R^s$. Assume $\E[U_Y|X,E]=0$ and
\begin{equation}
\label{eq:vector-tracer-example}
    Y\coloneqq\beta_0^\top X+\lambda_E^\top H+U_Y,
    \qquad\text{with}\qquad
    \E[H\mid X=x,E=e]=\E[H\mid X=x].
\end{equation}
Then $\mu_e(x)=\beta_0^\top x+\lambda_e^\top \E[H|X=x]$.
Fix any $e_0\in\{1,\ldots,J\}$ as a reference environment. Then the contrast from~\eqref{eq:contrast} is given by
\begin{equation}
\label{eq:vector-tracer-contrast}
\begin{aligned}
    C(x)%
    =L\,\E[H\mid X=x],\qquad \text{where} \qquad 
    L_{e,\cdot}&\coloneqq (\lambda_e-\lambda_{e_0})^\top,\qquad e\ne e_0.
\end{aligned}
\end{equation}
The contrast only needs to determine the baseline nuisance $\lambda_{e_0}^\top \E[H|X]$, not the full conditional expectation $\E[H|X]$. \Cref{thm:lowrank} gives sufficient conditions for identification in this setting. \Cref{prop:lowrank-general} in \Cref{app:proof-lowrank} extends the result to a general factorization of the conditional nuisances.
\begin{proposition}[Low-rank \MakeLowercase{\restrictionname}]
\label{thm:lowrank}
Suppose \eqref{eq:vector-tracer-example} holds with $\E[U_Y|X,E]=0$, and let~$L$ be defined by \eqref{eq:vector-tracer-contrast}. If
\[
\lambda_{e_0}\in\operatorname{rowspan}(L)
\qquad\text{and}\qquad
\E\!\left[\Var\!\left(X\mid \E[H\mid X],E\right)\right]\succ0,
\]
then \Cref{ass:cs,ass:pos} hold and $\beta_{\mathrm{\methodabbr}}=\beta_0$.
\end{proposition}

\subsection[Existence of at least one unconfounded environment]{
Existence of
at least one unconfounded environment}\label{sec:unconfounded}
\looseness-1 A simple sufficient condition arises when one environment has no
$X$-dependent bias. In that case, a contrast with this
environment directly reveals the nuisance in the remaining environments.
\begin{proposition}
\label{prop:clean}
Suppose some $e_\star\in \mathcal{E}$ satisfies $q_{e_\star}(X)=\kappa_\star$ almost surely for a constant $\kappa_\star$. If additionally $\Sigma_C\succ0$ (\Cref{ass:pos}), then 
$\beta_{\mathrm{\methodabbr}}=\beta_0$. 
\end{proposition}
Notably, knowledge of which environment is unconfounded is not required (so $e_\star$ can be unknown). Further, the assumption requires only that $q_{e_\star}(X)$ be constant; it does not require $H\indep X\mid E=e_\star$.

\subsection{Recoverable confounding state}
\label{sec:recoverable}
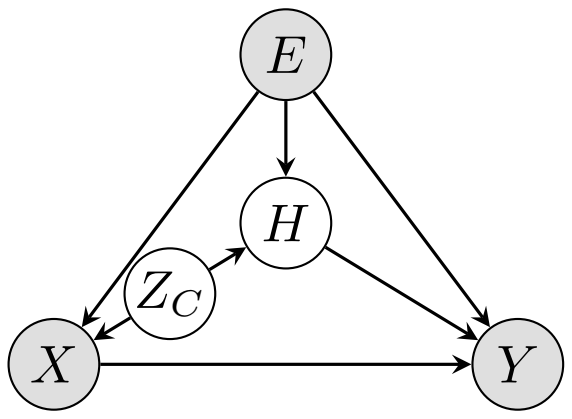
\begin{wrapfigure}[9]{r}{0.27\textwidth}
\captionsetup{font=footnotesize,skip=3pt,justification=centering}
\centering
\begin{tikzpicture}[
    node distance=5mm and 11mm,
    observed/.style={draw, circle, minimum size=6mm, inner sep=.25pt, , fill=gray!25},
    hidden/.style={draw, %
    circle, minimum size=6mm, inner sep=.25pt},
    arrow/.style={-{Latex[length=2mm]}, semithick}
]
    \node[hidden] (hid) {$H$};
    \node[observed, above=of hid] (env) {$E$};
    \node[observed, below left=of hid] (exp) {$X$};
    \node[observed, below right=of hid] (out) {$Y$};
    \path (hid) -- node[midway, hidden] (state) {$Z_C$} (exp);
    \draw[arrow, -stealth] (env) -- (exp);
    \draw[arrow, -stealth] (env) -- (hid);
    \draw[arrow, -stealth] (env) -- (out);
    \draw[arrow, -stealth] (state) -- (exp);
    \draw[arrow, -stealth] (state) -- (hid);
    \draw[arrow, -stealth] (hid) -- (out);
    \draw[arrow, -stealth] (exp) -- (out);
\end{tikzpicture}
\caption{Setting with recoverable latent-state $Z_C$.}
\label{fig:recoverable-state}
\vspace{.5em}
\end{wrapfigure}

In this setting, a latent state $Z_C$ (see~\Cref{fig:recoverable-state}) that can be recovered from the treatment~$X$  suffices to adjust for the confounding within each environment.

\begin{proposition}[Recoverable confounding state]
\label{prop:recoverable}
Suppose $Y\!\coloneqq\!\beta_0^\top X\!+\!\psi(H,E)\!+\!U_Y$ with ${\E[U_Y|X,E]\!=\!0}$. Suppose there are $Z_C\!\in\!\R^r$ and a measurable function $F$ such that $Z_C\!=\!F(X)$ almost surely and $X\!\indep\! H|(Z_C,E)$. For each $e\!\in\! \mathcal{E}$, 
define 
${\rho_e(z)\!\coloneqq\! \E[\psi(H,e)|Z_C\!=\!z,E\!=\!e]}$ 
and $\Delta(z)\!\coloneqq\! (\rho_e(z)\!-\!\rho_{e_0}(z))_{e\ne e_0}$. If $\Delta$ is injective on the support of $Z_C$ and ${\Sigma_{Z_C}\!\coloneqq\! \E[\Var(X|Z_C,E)]\succ0}$, then \Cref{ass:cs,ass:pos} hold, $\Sigma_C\!=\!\Sigma_{Z_C}$, and $\beta_{\mathrm{\methodabbr}}\!=\!\beta_0$.
\end{proposition}
\looseness-1
For a concrete example, consider an SCM with the graph in \Cref{fig:recoverable-state}. Here, the scalar state $Z_C>0$ is the treatment vector's Euclidean length: $Z_C=\|X\|$
almost surely. Write $X=Z_CV\in\R^d$, where $V\in\R^d$ is its direction and $\|V\|=1$ almost surely.
Let the (scalar) confounder
$H$ and outcome $Y$ satisfy 
$
H\coloneqq Z_C^2+b_E Z_C+ U_H,
$ and $
Y\coloneqq\beta_0^\top X+H+d_E+U_Y,
$
where $U_H$ and $U_Y$ are mutually independent, centered scalar noises, independent of $(Z_C,V,E)$; and $(b_e,d_e)_{e\in\mathcal{E}}$ are environment-specific constants.

For %
fixed length~$Z_C$ and environment $E$, the direction $V$ provides no additional information about $H$, so ${X\!\indep\! H|(Z_C,E)}$.
Here, ${\rho_e(z)\!=\!z^2+b_ez+d_e}$ and ${q_e(X)\!=\!\rho_e(Z_C)}$.
Hence, ${C(X)\!=\!\Delta(Z_C)}$, where
$
\Delta(z)
=\big((b_e-b_{e_0})z+(d_e-d_{e_0})\big)_{e\ne e_0}.
$
If $b_e\ne b_{e_0}$ for at least one environment, then $\Delta$
is injective. If additionally
${\E[Z_C^2\Var(V|Z_C,E)]\succ0}$, \Cref{prop:recoverable}
implies $\beta_{\mathrm{\methodabbr}}=\beta_0$.
To apply \methodabbr{}, we use $C(X)$ and $E$ as adjustment
variables; we need not specify that $Z_C=\|X\|$.

\section{Discussion in the context of related work}
\label{sec:discussion}
\looseness-1 Several methods identify coefficients of linear structural equation models using changes in additive noise distributions across environments, while keeping the structural coefficients fixed.
Causal Dantzig~\citep{rothenhausler2019causal} requires $\E[X(Y-\beta_0^\top X)|E=e]$ to be the same in every environment and estimates $\beta_0$ by differencing these moment equations.
KL regression~\citep{gimenez2020kl} fits a common causal parameter
to the covariance matrices of~$(X,Y)$ across environments; it allows the noise covariance of $X$ and noise variance of $Y$ to change while keeping the confounder distribution and its effects fixed.
BackShift~\citep{rothenhausler2015backshift} estimates the structural coefficients among all observed variables (allowing feedback loops) from covariance differences. It assumes an invariant baseline noise distribution and additive shifts that are uncorrelated across variables and with the baseline noise.
Our setting allows the effect of $H$ on $Y$ to change across
environments while $\beta_0$ remains fixed. Such changes can
violate Causal Dantzig's moment invariance and the invariant
confounding structure assumed by KL regression and BackShift,
so their identification guarantees do not cover our setting
in general.
\methodabbr{} forms contrasts of the conditional expectations $\mu_e(x)$ at fixed $x$ and adjusts for these contrasts under \restrictionname{} and positivity.

\looseness-1 Perhaps the most closely related result to our work is Thm.~3.4 of \citet{kivva2025causal}, which identifies a scalar effect when one latent-outcome loading changes across two environments, using a linear independent-noise model and higher-moment or non-Gaussianity conditions. In their loading-shift setting, \restrictionabbr{} holds because the contrast is a nonzero multiple of $\E[H|X]$; positivity, 
however,
fails if this conditional expectation is injective in $X$. Our identification result allows vector treatments and nonlinear conditional nuisances under \restrictionname{} and positivity. It requires finite second moments but no parametric noise distribution or higher-order-moment identifying equations.

\looseness-1 Unlike invariance-based and
adjustment-set methods~\citep{peters2016causal,arjovsky2019invariant,
debartolomeis2025doubly}, \methodabbr{} does not seek an invariant predictor nor an observed adjustment set.

\looseness-1
Environment IV~\citep{newey1999nonparametric}, anchor regression~\citep{rothenhausler2021anchor}, and control-function approaches~\citep{gnecco2023boosted} exploit environment-induced variation in the treatment $X$, with method-specific exclusion or stability conditions
\methodabbr{} instead uses variation in the nuisance~$q_e(x)$: the \restrictionname{} requires $C(X)$ to determine~$q_{e_0}(X)$, and positivity requires residual variation in $X$ given $(C(X),E)$.
Under these assumptions, \methodabbr{} can 
identify $\beta_0$ without an $E\to X$ shift. 
Conversely, invariant confounding terms cancel from the contrast and thus can generally not be handled by OVR, 
but 
a valid environment instrument may still identify $\beta_0$ in such a setting.

\looseness-1 
Proxy-variable 
methods use additional observed variables that carry information about $H$, together with proxy-exclusion restrictions and bridge or completeness conditions \citep{miao2018identifying,tchetgen2024introduction}. \methodabbr{} uses a different object: $C(X)$ is generated from environment-specific conditional expectations and is a deterministic function of $X$, not an additional measurement of $H$. \restrictionabbr{} requires $C(X)$ to determine $q_{e_0}(X)$; it does not require recovering $H$ or solving a bridge equation. The approaches are therefore complementary. Proxy methods 
do not require
environment variation when suitable proxies are observed, whereas \methodabbr{} 
does not require a
measured proxy but requires informative outcome-mechanism shifts and positivity.
\Cref{app:related} 
discusses further related work.

\section{Experiments}
\label{sec:experiments}
\label{sec:estimators}
We use Smoothed \methodabbr{} (\Cref{alg:smoothed-ccr}) with smoothing scale $\tau=1$.
For each observation, we average predictions at $M=8$ perturbed contrast inputs, using four independent Gaussian noise draws and their negatives. \Cref{app:dgps} gives implementation and experimental details. \Cref{app:ablation} compares Direct 
(based on simple plug-in 
estimation)
and Smoothed \methodabbr{}. 
Code for reproducing the experiments is hosted at \url{https://github.com/fschur/outstrumental-variable-regression}.

\subsection{Synthetic study}
\label{sec:synthetic_experiments}
We compare Smoothed \methodabbr{} (\Cref{alg:smoothed-ccr}) with linear regression of $Y$ on $X$ (OLS), linear environment fixed-effects regression (FE), 
a jointly trained linear-plus-neural-network model (LNN),
which fits ${Y\approx\beta^\top X+f_\theta(X,E)}$ by squared error (\Cref{app:lnn}), and anchor regression~\citep{rothenhausler2021anchor} with $\gamma=2$.
The estimator of \citet{kivva2025causal} assumes a scalar treatment and a linear independent-noise SCM and is therefore not directly applicable to our setting, where treatments are multivariate and the confounding mechanisms need not be linear.

We consider several different data generating processes.
\begin{figure}[t]
\centering
\resizebox{\linewidth}{!}{\input{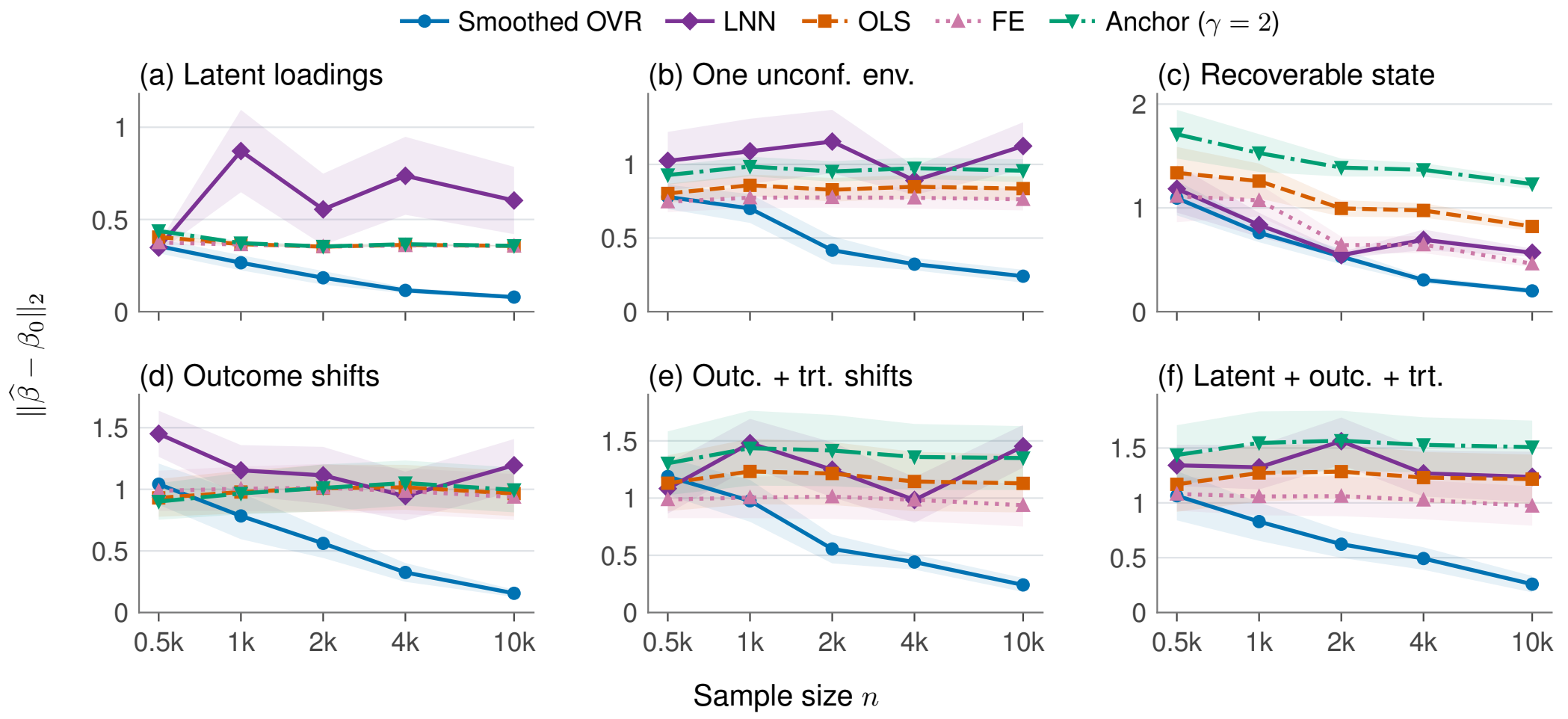}}
\caption{Coefficient error at $n\in\{0.5,1,2,4,10\}\mathrm{k}$ across six synthetic settings. Lines show means and bands $\pm$ one Monte Carlo s.e.\ over ten paired replications; panels use separate linear error scales. As $n$ increases, Smoothed OVR has the lowest mean $\ell_2$ estimation error in all six settings.
}
\label{fig:synthetic-core}
\end{figure}
\Cref{fig:synthetic-core} reports the means $\smash{\|\widehat\beta-\beta_0\|_2}$ and standard errors for different estimators over 
ten replications, using the same generated observations for all methods within each replication, at sample sizes $n\in\{500,1000,2000,4000,10000\}$.
Panels (a)--(c) correspond to the three sufficient settings in \Cref{sec:summary}. Positivity is not guaranteed in setting (b). Panels (d)--(f) do not impose \restrictionabbr{} and progressively add shifts in the outcome mechanism, the treatment distribution, and the latent-confounder distribution.
The precise SCMs and implementation details are specified in \Cref{app:dgps}.

\textbf{(a) Latent loadings (\cref{sec:lowrank}).}
Two binary confounders $H_1,H_2$ affect disjoint pairs of treatment components: $H_1$ affects $(X_1,X_2)$ and $H_2$ affects $(X_3,X_4)$.
The environment $E$ changes only their coefficients in the outcome equation.
The loading-difference matrix $L$ has full column rank, so
$\E[(H_1,H_2)^\top|X]$ can be recovered from $C(X)$.
Variation within each pair ensures positivity, satisfying the conditions of \Cref{thm:lowrank}.

\textbf{(b) One unconfounded environment (\cref{sec:unconfounded}).}
The environment $E$ induces additive shifts in the equations for
$H$ and $X$, with $X$ depending nonlinearly on $H$.
One environment (it is unknown to the method which one) is unconfounded ($q_{e_\star}=0$). The contrast with this environment reveals the reference nuisance as in \Cref{prop:clean}, while Gaussian treatment noise gives common support.

\textbf{(c) Recoverable confounding state (\cref{sec:recoverable}).}
We use the special case after \Cref{prop:recoverable} with ${Z_C=\|X\|}$:
within each environment, $q_e(X)$ depends only on this length.
After adjustment, variation in the direction of $X$ leaves positive
residual variance in every nonzero linear combination of~$X$,
ensuring positivity.

\textbf{(d) Outcome shifts.}
Here $E\indep(H,X)$, and $Y$ is the only child of $E$, as in~\cref{fig:ovr-setting} without the dashed edges.
A non-linear function maps a scalar confounder $H$ and independent
noise to $X\in\mathbb R^4$.
The contribution of $H$ to $Y$ is represented by a separate
neural network in each environment, with weights drawn
independently across environments.
We do not impose the structure used in \Cref{sec:summary} to ensure \restrictionabbr{}, so $q_{e_0}(X)$ is generally not perfectly recoverable
from $C(X)$.

\textbf{(e) Outcome and treatment shifts.}
Starting from (d), we add an environment-specific mean shift to $X$ (that is, $E\to X$ in \cref{fig:ovr-setting}). Thus $P(H|X,E)$ can now depend on $E$,
even though $E\indep H$ still holds. As in (d), we impose neither \restrictionabbr{} nor positivity.

\textbf{(f) Outcome, treatment, and confounder shifts.} 
Starting from (e), we also add an environment-specific mean shift to $H$
(that is, $E\to H$ in \cref{fig:ovr-setting}). Thus $E$ now directly
affects $H$, $X$, and $Y$. As in (d) and (e), we impose neither
\restrictionabbr{} nor positivity.

\Cref{fig:synthetic-core} shows that the mean $\ell_2$ estimation error of Smoothed \methodabbr{} decreases with $n$ in all six settings. Smoothed \methodabbr{} is competitive at smaller sample sizes and has the lowest mean error in every setting for sample sizes $n> 2{,}000$. 
Panels (d)--(f) impose neither \restrictionabbr{} nor positivity, so identification is not guaranteed by construction. The contrasts may nevertheless remove much of the confounding;  the bound in \cref{sec:sensitivity} shows that population bias can remain small when violations of \restrictionabbr{} are small and sufficient treatment variation remains after adjustment.

\begin{table}[t]
\centering
\caption{
Mean $\ell_2$ estimation error over ten training/reference splits, with Monte Carlo standard errors in parentheses and the lowest means in bold.
}
\label{tab:real-data}
\small
\setlength{\tabcolsep}{4pt}
{
\begin{tabular}{@{}ccccc@{}}
\toprule
Smoothed \methodabbr & OLS & FE & Anchor ($\gamma=2$) & LNN \\
\midrule
$\mathbf{.019}\;(\mathbf{.001})$ & $.129\;(.001)$ & $.104\;(.001)$ & $.171\;(.001)$ & $.095\;(.011)$ \\
\bottomrule
\end{tabular}
}
\end{table}
\subsection{Real-data benchmark}
\label{sec:real-data}

\looseness-1 We use KuaiRand-Pure \citep{gao2022kuairand}, a dataset of user-video interactions from the Kuaishou short-video platform. Videos are selected either by the platform's usual recommender system (standard policy) or randomly from a candidate pool (random policy). We use these two assignment policies as the environments $E$.
We retain users and videos observed under both policies and sample equal policy counts within each user. The resulting cohort has $339{,}954$ interactions, $20{,}669$ users, and $5{,}613$ videos. The standardized treatment vector $X\in\R%
^6$ contains log video duration, log upload age, log aspect ratio, indicators for long-import and web uploads, and tag count. The outcome is ${Y \mathrel{\coloneqq}\log(1+\min\{\text{watch seconds},300\})}$.

For each of ten repetitions, we split interactions separately for each user and policy. The resulting two training environments contain $80{,}079$ standard and $80{,}079$ random interactions; all methods use these same $160{,}158$ observations. 
In each repetition, we fit a separate linear regression of~$Y$ on~$X$
using the remaining $89{,}898$ random-policy interactions.
Its fitted coefficients serve as the reference for evaluation. 
No observation appears in both the training and reference sets. The same $20{,}669$ users appear in both the training and reference sets
to keep the target population fixed. The random policy breaks assignment based on the recommender's user state, but does not independently randomize the six video attributes or eliminate unmeasured item characteristics. 
We evaluate distance to the reference linear-regression coefficients
(computed in the reference sets with random video assignment); these are not
estimates of easily interpretable structural coefficients.
Crucially, none of the methods are told which of the two training environments corresponds to the reference random policy.

We evaluate all methods by their mean $\ell_2$ estimation error to the reference coefficients, see~\Cref{tab:real-data} for the results.
Smoothed \methodabbr{} achieves the lowest mean $\ell_2$ estimation error over ten training/reference splits ($.019$), compared with $.095$ for LNN, the closest baseline.

\section{Conclusion and future work}
\label{sec:outlook}
\looseness-1 
\methodname{} complements methods that require environment-induced treatment shifts such as IV by turning outcome heterogeneity into an adjustment signal.
We establish identifiability under the \MakeLowercase{\restrictionname} and positivity, and consistency of the smoothed estimator under additional regularity conditions.
Future work includes extensions to non-linear effects, e.g.\ via feature maps, and to continuous environments, where the contrast becomes function-valued.

\bibliographystyle{iclr2027_conference}
\bibliography{iclr2027_conference}

\appendix

\section{Extended Related Work}
\label{app:related}

\subsection{Multi-environment identifiability and invariance}

Causal Dantzig identifies a linear effect from cross-environment changes in equations involving $\E[XY]$ and $\E[XX^\top]$ under additive interventions \citep{rothenhausler2019causal}, while KL regression fits a common linear-SEM coefficient to environment-specific joint covariances under shift interventions \citep{gimenez2020kl}. BackShift targets the entire connectivity matrix of a linear, potentially cyclic equilibrium SEM by jointly diagonalizing covariance differences \citep{rothenhausler2015backshift}. It permits latent confounding through a non-diagonal noise covariance, but requires both that this covariance and the structural matrix remain invariant and that the unknown intervention shifts be uncorrelated with the structural noise and have diagonal covariance. A changing latent-outcome loading---including our motivating outcome-only mechanism shift---violates these invariance conditions; moreover, BackShift's identifiability result for the full structural coefficient matrix needs at least three environments, whereas \methodname{} can identify $\beta_0$ with two. These methods exploit changes in second moments under invariant linear structural equations, while \methodname{} permits the conditional confounding nuisance itself to vary and can identify $\beta_0$ with $P(X\mid E)$ unchanged.

The closest result also using variation in the outcome to identify $\beta_0$ is Theorem~3.4 of \citet{kivva2025causal}, which also allows an unchanged treatment distribution when a latent-outcome loading changes, but assumes a scalar linear independent-noise graph and uses higher-moment or non-Gaussianity conditions. \methodname{} differences the full conditional expectations, permitting vector treatments and nonlinear conditional nuisances under an invariant linear coefficient in a partially linear model, \restrictionabbr, and positivity.

RAMEN searches for an observed adjustment set through conditional-mean invariance and requires all parents of either $X$ or $Y$ to be observed and that node's conditional mean to be invariant across environments \citep{debartolomeis2025doubly}; \methodname{} instead uses outcome mechanisms that vary across environments and adjusts for the contrast $C(X)$ to identify $\beta_0$. \citet{karlsson2023detecting} test for hidden confounding between treatment and outcome after adjustment for observed covariates, using data from multiple environments. \citet{mameche2024identifying} use dependent changes in observed conditional distributions to identify groups of variables with a common hidden cause. These methods target the presence and structure of hidden confounding, whereas \methodname{} identifies the causal parameter $\beta_0$ under \restrictionabbr{} and positivity. Their detection guarantees impose assumptions on how mechanisms change, including independence across mechanism changes; these assumptions do not follow from \restrictionabbr. Invariant causal prediction tests observed covariate subsets for an environment-invariant conditional outcome law with the goal of identifying the causal parents of $Y$ \citep{peters2016causal}, whereas invariant risk minimization learns representations supporting a common risk-optimal predictor for out-of-distribution prediction \citep{arjovsky2019invariant}. \methodname{} uses the opposite signal: outcome non-invariance reveals the conditional nuisance while the causal effect remains invariant.

\looseness-1 
The broader idea that sufficiently rich %
environmental variation can
reveal otherwise latent structure also appears in nonlinear independent
component analysis with auxiliary variables
\citep{hyvarinen2019nonlinear,khemakhem2020variational} and %
interventional
causal representation learning~\citep{von2023nonparametric,wendong2023causal,varici2025score}. In
these settings, identifiability likewise relies on sufficiently diverse
changes across environments, although the objects being contrasted and the
resulting identification problems are different.

\subsection{Instruments and control functions}

Environment IV methods use $E$ as an instrument, as depicted in \Cref{fig:iv-setting}, and extend to time series \citep{thams2024identifying, tiedemann2024identifying}, while classical control functions construct a control from a treatment first stage under triangular-model exclusion and stability restrictions \citep{newey1999nonparametric}.
Anchor regression approaches an IV/2SLS-type estimand as its robustness penalty grows; that limit identifies the causal parameter only under the corresponding instrument-validity conditions \citep{rothenhausler2021anchor}. Under a simultaneous-equation shift model and sufficient heterogeneity, boosted control functions instead target an invariant worst-case predictor that need not equal the structural response \citep{gnecco2023boosted}. \methodname{} does not use $E$ as an instrument: it allows a direct environment-dependent outcome nuisance and can operate when $E\indep X$. Conversely, under their respective identification assumptions, IV and covariance-based methods can identify $\beta_0$ in settings in which \methodname{}'s positivity condition fails.

\subsection{Proxy-based deconfounding and semiparametric estimation}

Proxy methods use observed variables that carry information about unobserved confounders to identify causal effects under additional assumptions.
The deconfounder infers a substitute confounder from a multi-cause assignment model \citep{wang2019blessings}, although multiple causes alone do not generally identify causal effects \citep{damour2019multi}; causally supervised text embeddings instead compress an observed variable assumed sufficient for adjustment \citep{veitch2020adapting}. Proximal methods begin with measured treatment- and outcome-side proxies and impose proxy exclusions, bridge existence, and completeness conditions \citep{miao2018identifying,tchetgen2024introduction}. Proxy information can also enable the transfer of causal effects across
domains under unobserved confounding
\citep{iglesias2026transferring}.
Single-proxy identifiability instead relies on a negative-control outcome or a known proxy mechanism plus completeness \citep{park2024single,vollmer2026identifying}. The contrast $C(X)$ is different from the proxies considered by the other works cited in this paragraph: it is generated from environment-specific conditional expectations and is a deterministic function of $X$, rather than a measured proxy for $H$ or a proximal bridge. \restrictionabbr{} requires $q_{e_0}(X)$ to be a measurable function of $C(X)$, not to recover the latent confounder itself.

\subsection{Alternative structure for unobserved confounding}
A different line of work obtains causal identification by exploiting
structure in the unobserved heterogeneity rather than exclusion restrictions
or measured proxies. In panel settings, synthetic-control and
interactive-fixed-effect methods use cross-unit and temporal structure
that can be represented through latent factor models, while matrix-completion
approaches impose related low-rank structure on untreated potential outcomes
\citep{abadie2010synthetic,xu2017generalized,athey2021matrix}.
For time-series data, \citet{schur2026decor} instead exploit spectral
sparsity of an unobserved confounding process and reduce causal effect
estimation to robust regression in the frequency domain. These approaches
are complementary to \textsc{\methodabbr}: they rely on structural restrictions
across units or time, whereas \textsc{\methodabbr} exploits informative variation
of the outcome mechanism across environments.

\subsection{Differencing across data sources to expose confounding}
\label{app:differencing}

More broadly, the elementary step behind \textsc{\methodabbr}, differencing quantities across which the
causal component is invariant so that only the confounding component remains, also
underlies several established approaches. Difference-in-differences compares outcome
changes between treated and untreated groups under parallel trends, that is, an
additive, covariate-invariant confounding shift
\citep{card1994minimum,roth2023whats}. Negative controls and control-outcome
calibration use an auxiliary outcome known to be unaffected by the treatment, whose
observed association with the treatment calibrates the confounding bias in the primary
outcome \citep{lipsitch2010negative,tchetgen2014control}; the single-proxy methods
discussed above build on the same idea. Experimental grounding fits a flexible effect
model on confounded observational data and corrects it by a parametric function of the
difference to a small unconfounded trial \citep{kallus2018removing}. More generally,
methods that combine randomized and observational data assume that one source is
unconfounded, or observes the full confounder set, and use the other source for
efficiency or extrapolation
\citep{yang2020combining,hatt2022combining,rosenman2023combining,yang2023elastic};
see \citet{colnet2024causal} for a review.

\section{Estimation Details}
\label{app:estimation}

This section explains the direct estimator, the contrast-network implementation, and the residual-variation diagnostic used later on in the appendix.

\textbf{Direct estimator.} Direct \methodabbr{} (D\methodabbr; \Cref{alg:direct-ccr}) implements the residual regression in \eqref{eq:population-tracer} using estimated contrasts without smoothing. We abbreviate Smoothed \methodabbr{} (\Cref{alg:smoothed-ccr}) as S\methodabbr.

\begin{algorithm}[H]
\caption{Direct \methodname}
\label{alg:direct-ccr}
\small
\textbf{Input:} Data $\{(X_i,Y_i,E_i)\}_{i=1}^n$, reference $e_0$, folds $K$.
\begin{enumerate}
    \item Split the data into $\mathcal I_C$ and $\mathcal I_\beta$. On
    $\mathcal I_C$, estimate $\widehat\mu_e$ and set
    \[
        \widehat C(x)
        \coloneqq 
        \bigl(\widehat\mu_e(x)-\widehat\mu_{e_0}(x)\bigr)_{e\ne e_0}.
    \]

    \item Cross-fit regressions of $X$ and $Y$ on
    $(\widehat C(X),E)$ within $\mathcal I_\beta$, yielding held-out residuals
    $(\widehat X_i,\widehat Y_i)$.

    \item Return
    \[
        \widehat\beta_{\mathrm{direct}}
        \mathrel{\coloneqq}
        \left(\sum_{i\in\mathcal I_\beta}
        \widehat X_i\widehat X_i^\top\right)^{-1}
        \sum_{i\in\mathcal I_\beta}\widehat X_i\widehat Y_i.
    \]
\end{enumerate}
\end{algorithm}

\textbf{Contrast estimation.} Our implementation of the first step of D\methodabbr{} fits $\mu_e(x)=f(x)+c_e(x)$ jointly across environments, with a shared component $f$ and contrast heads satisfying $c_{e_0}\equiv0$. A shared network maps $x$ to features $h_\theta(x)$. Separate linear output layers give $f(x)=w_0^\top h_\theta(x)+b_0$ and $c_e(x)=w_e^\top h_\theta(x)+b_e$ for $e\ne e_0$; we call the latter layers contrast heads. The fitted contrast is $(\widehat c_e(x))_{e\ne e_0}$. Joint fitting helps prevent the invariant signal from leaking into separately estimated contrasts. The smoothed procedure in \Cref{alg:smoothed-ccr} instead uses two contrast fits and noisy-input residualization; \Cref{thm:consistency-full} establishes consistency of S\methodabbr.

\subsection[Positivity diagnostic]{ Positivity diagnostic }
\label{sec:diagnostics}

Let $\widehat X_i$ be the held-out treatment residuals, using averaged predictions of $X$ at perturbed contrast inputs as in Smoothed \methodabbr{}, and set $\smash[tb]{\widehat\Sigma
\mathrel{\coloneqq}|\mathcal I_\beta|^{-1}
\sum_{i\in\mathcal I_\beta}\widehat X_i\widehat X_i^\top}$. With $\widehat\Sigma_E$ the covariance after residualizing $X$ only on $E$, assumed positive definite, we report the positivity diagnostic $\widehat\gamma\mathrel{\coloneqq}\lambda_{\min}(\widehat\Sigma_E^{-1/2}\widehat\Sigma\widehat\Sigma_E^{-1/2})$, which equals
\[
 \widehat\gamma=\min_{v\ne0}\frac{v^\top\widehat\Sigma v}{v^\top\widehat\Sigma_Ev}.
\]
Thus it compares the fitted residual second moment with within-environment variation in the least-retained linear direction. The ratio is invariant under invertible linear changes of treatment coordinates applied to both covariance matrices. For the true contrast and exact conditional expectations, the population ratio uses $\Sigma_C$ and $\Sigma_E\coloneqq\E[\Var(X\mid E)]$; conditional total variance gives $0\preceq\Sigma_C\preceq\Sigma_E$, so that ratio lies in $[0,1]$. Estimated conditional means need not satisfy this bound. Small $\widehat\gamma$ flags little fitted residual variation relative to within-environment variation, without a universal cutoff. This diagnostic measures variation in the fitted residuals and does not by itself verify population positivity (\Cref{ass:pos}).

\section[Consistency of Smoothed \methodabbr]{Consistency of Smoothed \methodabbr}
\label{app:theory}
{
\textbf{Sampling and notation.}
In this section, we use notation defined in \Cref{alg:smoothed-ccr}. The fitted functions $f_{n,k,Z}$ correspond to $\widehat f_{Z,k}$ in the algorithm; $m_{n,k,Z}$ is the corresponding population conditional expectation given the perturbed contrast, the environment, and the training data.
We observe an i.i.d.\ sample from $P$ as in \Cref{sec:problem_setting}. Define $W\coloneqq Y-\mu_E(X)$. Then $Y=\beta_0^\top X+q_E(X)+W$ and $\E[W\mid X,E]=0$.
Under \Cref{ass:cs}, $q_E(X)=\varphi_E(C(X))$ with $\varphi_e(c)\coloneqq\varphi(c)+c_e$ and $c_{e_0}\coloneqq0$.
We analyze \Cref{alg:smoothed-ccr} with splits drawn independently of the observations, $|\mathcal I_C|=n_C\to\infty$, $|\mathcal I_\beta|=n\to\infty$, and a fixed number $K$ of folds with $|I_k|\to\infty$. Environment-stratified splits are also covered if every split and fold has limiting environment proportions $P(E=e)$; the laws of large numbers below are then applied within each environment.
Write $\widehat C_n\coloneqq\overline C$, $V_{n,k}\coloneqq\tau^2\widehat\Omega_k$ for fixed $\tau$, and $f_{n,k,Z}\coloneqq\widehat f_{Z,k}$ for $Z\in\{X,Y\}$.

Let $\mathcal T_{n,k}$ be the collection of the first-stage fits, the training observations outside $I_k$, and the nuisance-fitting randomness. For a fresh observation independent of $\mathcal T_{n,k}$, let $U_{n,k}\mathrel{\coloneqq}\widehat C_n(X)+\xi$, where $\xi\mid\mathcal T_{n,k}\sim N(0,V_{n,k})$ is independent of the observation, and define
\[
 m_{n,k,Z}(u,e)\mathrel{\coloneqq}\E[Z\mid U_{n,k}=u,E=e,\mathcal T_{n,k}],
 \qquad Z\in\{X,Y,q_E(X)\}.
\]
\Cref{thm:consistency-full,lem:smoothing-overlap} cover any perturbation scheme in which each held-out row uses $M$ perturbations $\xi_{i1},\ldots,\xi_{iM}$ with marginals $N(0,V_{n,k})$, shared by the $X$ and $Y$ predictions, independent across rows and of the held-out observations given $\mathcal T_{n,k}$; the antithetic pairs of \Cref{alg:smoothed-ccr} are a special case. Only the marginal laws enter the argument; the perturbations within a row may be independent or form antithetic pairs. Allowing dependence therefore covers the implementation in \Cref{alg:smoothed-ccr}.
For $i\in I_k$ and $Z\in\{X,Y\}$, the fitted residuals are
\[
\widehat Z_i\coloneqq Z_i-\frac1M\sum_{j=1}^M
f_{n,k,Z}\bigl(\widehat C_n(X_i)+\xi_{ij},E_i\bigr).
\]
Let $\widehat G_n\mathrel{\coloneqq}n^{-1}\sum_{i\in\mathcal I_\beta}\widehat X_i\widehat X_i^\top$ and
$\widehat\beta_{\mathrm{SOVR}}\mathrel{\coloneqq}(\widehat G_n+\lambda_n I)^{-1}n^{-1}\sum_{i\in\mathcal I_\beta}\widehat X_i\widehat Y_i$, where $\lambda_n\ge0$ tends to zero; $\lambda_n=0$ gives the estimator of \Cref{alg:smoothed-ccr}, set arbitrarily if $\widehat G_n$ is singular. Thus \Cref{alg:smoothed-ccr} is covered with $\lambda_n=0$. A positive $\lambda_n$ ensures that the sample matrix is invertible at every sample size; requiring $\lambda_n\to0$ makes the ridge term vanish in the limit.

\begin{assumption}[Smoothed consistency regularity]
\label{ass:est}
For every fold $k$:
\begin{enumerate}[label=(\alph*),leftmargin=*,nosep]
\item The contrast error satisfies
$\rho_n\mathrel{\coloneqq}\|\widehat C_n-C\|_{L^2(P_X)}\to_P0$, and $\operatorname{tr}(V_{n,k})\to_P0$.
\item With $\delta_n(X)\mathrel{\coloneqq}\widehat C_n(X)-C(X)$ and
$\zeta_{n,k}\mathrel{\coloneqq}\delta_n(X)-\E[\delta_n(X)\mid C(X),E,\mathcal T_{n,k}]$, $\zeta_{n,k}$ satisfies
\[
 \zeta_{n,k}\in\operatorname{range}(V_{n,k})\ \text{a.s.},\qquad
 \kappa_{n,k}\mathrel{\coloneqq}\E[\zeta_{n,k}^\top V_{n,k}^{\dagger}\zeta_{n,k}\mid\mathcal T_{n,k}]=O_P(1).
\]
Here $V^\dagger$ is the Moore--Penrose inverse.
\item For $Z=X,Y$, the noisy-input regression error satisfies
\[
 \E\!\left[\|f_{n,k,Z}(U_{n,k},E)-m_{n,k,Z}(U_{n,k},E)\|^2
                \mid\mathcal T_{n,k}\right]=o_P(1).
\]
\end{enumerate}
\end{assumption}

We call $\zeta_{n,k}$ the leakage component: it is the contrast error after subtracting its conditional mean given $(C,E,\mathcal T_{n,k})$. Part (b) controls this within-$(C,E)$ variation relative to the perturbation covariance. The range condition excludes unperturbed leakage when $V_{n,k}$ is singular. \Cref{app:finite-feature-calibration} gives a sufficient condition for parts (a) and (b).

The following theorem is the formal version of \Cref{thm:main-consistency}.
\begin{theorem}[Consistency of Smoothed \methodabbr]
\label{thm:consistency-full}
Under the sampling scheme above and \Cref{ass:cs,ass:pos,ass:est}, for every fixed number of perturbations $M$ with $\xi_{ij}\mid\mathcal T_{n,k}\sim N(0,\tau^2\widehat\Omega_k)$ for every $i\in I_k$ and $j\in\{1,\ldots,M\}$,
\[
 P(\widehat G_n\succ0)\longrightarrow1,\qquad
 \|\widehat G_n^{-1}\|=O_P(1),\qquad
 \widehat\beta_{\mathrm{SOVR}}\xrightarrow{p}\beta_0.
\]
\end{theorem}

\subsection[Sufficient condition for Assumption~\ref*{ass:est}(a) and (b)]{Sufficient condition for \Cref{ass:est}\textnormal{(a)} and \textnormal{(b)}}
\label{app:finite-feature-calibration}
\begingroup

\Cref{ass:est}(a) and (b) require vanishing perturbations whose covariance controls the leakage component of the estimated contrast. \Cref{prop:finite-feature-calibration} proves that both requirements hold for correctly specified regression on fixed features: disagreement between two independent fits provides a covariance of the required order. It does not establish the noisy-input regression condition in \Cref{ass:est}(c).
\begin{proposition}
\label{prop:finite-feature-calibration}
Assume the sampling assumptions stated in the ``Sampling and notation'' paragraph of \Cref{app:theory}.
Let $d_C\coloneqq\dim C=|\mathcal E|-1$ and let $\phi(X)\in\R^p$, with $p\ge d_C$, satisfy $\E\|\phi(X)\|^2<\infty$ and $\Gamma\mathrel{\coloneqq}\Cov(\phi(X))\succ0$. Suppose
\[
C(x)=B_0\phi(x),\qquad
\widehat C^s(x)=\widehat B^s\phi(x),\quad s\in\{A,B\},
\]
where $B_0,\widehat B^A,\widehat B^B\in\R^{d_C\times p}$. Assume the two fits, based on disjoint halves of the contrast sample, satisfy
\begin{equation}
\label{eq:calibration-gaussian-limit}
\sqrt{n_C}\bigl(\widehat B^A-B_0,\widehat B^B-B_0\bigr)
\ \xrightarrow{d}\ (G_A,G_B)\qquad\text{as }n_C\to\infty,
\end{equation}
where $G_A$ and $G_B$ are independent Gaussian matrices with nonsingular covariance matrices after vectorization.

For each fold $k$, let $\widehat\Gamma_{n,k}$ be the empirical covariance of $\phi(X)$ on $\mathcal I_\beta\setminus I_k$. Also assume that $\widehat B^A-\widehat B^B$ has row rank $d_C$ and $\widehat\Gamma_{n,k}\succ0$ almost surely for all sufficiently large sample sizes.

Then, for every fixed $\tau>0$, \Cref{ass:est}(a) and (b) hold.
\end{proposition}

\begin{proof}
Fix $\tau>0$. Under the fixed-feature representation, the disagreement covariance satisfies
\[
V_{n,k}=\frac{\tau^2}{4}
(\widehat B^A-\widehat B^B)\widehat\Gamma_{n,k}
(\widehat B^A-\widehat B^B)^\top.
\]
The sampling assumptions give $\widehat\Gamma_{n,k}\to_P\Gamma$, so
\[
n_CV_{n,k}\ \Rightarrow\
\frac{\tau^2}{4}(G_A-G_B)\Gamma(G_A-G_B)^\top.
\]
The limiting matrix is positive definite almost surely: $p\ge d_C$ and the Gaussian coefficient difference has a density on $\R^{d_C\times p}$. Hence $\lambda_{\min}(V_{n,k})^{-1}=O_P(n_C)$. The coefficient expansion and finite feature second moments also give $\rho_n^2=O_P(n_C^{-1})$. Since $V_{n,k}$ is full rank, the range condition holds, and conditional projection yields
\[
\kappa_{n,k}
\le \frac{\E[\|\zeta_{n,k}\|^2\mid\mathcal T_{n,k}]}{\lambda_{\min}(V_{n,k})}
\le \frac{\rho_n^2}{\lambda_{\min}(V_{n,k})}
=O_P(1).
\]
Thus \Cref{ass:est}(b) holds. Moreover, $\operatorname{tr}(V_{n,k})=O_P(n_C^{-1})$, so part (a) holds as well.
\end{proof}

The Gaussian limit assumption~\eqref{eq:calibration-gaussian-limit} is the usual asymptotic normality condition for correctly specified fixed-dimensional regression with nondegenerate coefficient estimation error.

We regard this calibration argument as informative even when some assumptions may not hold; for example, the parametric convergence rates assumed here are not established for the neural-network estimators used in \Cref{app:estimators}.
\par
\endgroup

\subsection[Gaussian overlap and residual variation]{Gaussian overlap and residual variation}
\label{app:smoothing-overlap}
We use the following lemma in the proof of \Cref{thm:consistency-full}. Under \Cref{ass:pos,ass:est}(b), it bounds the residual Gram matrix away from singularity in probability, including when predictions are averaged over antithetic perturbations.
\begin{lemma}[Preservation of residual variation]
\label{lem:smoothing-overlap}
Fix a fold and condition on $\mathcal T_{n,k}$, suppressing $k$ in the notation. Under \Cref{ass:pos,ass:est}(b), let
$Q_n\mathrel{\coloneqq}\E[\Var(X\mid U_n,E,\mathcal T_{n,k})\mid\mathcal T_{n,k}]$.
Draw $X,X'$ independently conditional on the same $(C,E)$, and put
\[
 \begin{aligned}
 \Delta X&\mathrel{\coloneqq}X-X',\qquad
 d_n^2\mathrel{\coloneqq}\{\widehat C_n(X)-\widehat C_n(X')\}^\top
       V_n^\dagger\{\widehat C_n(X)-\widehat C_n(X')\},\\
 B_n&\mathrel{\coloneqq}\tfrac12\E[\Delta X\Delta X^\top e^{-d_n^2/4}\mid\mathcal T_{n,k}].
 \end{aligned}
\]
Then $Q_n\succeq B_n$ and $\lambda_{\min}(B_n)^{-1}=O_P(1)$.
For the average $\widetilde m_{n,X}$ of the conditional expectations $m_{n,X}$ over the $M$ shared perturbations, the residual Gram matrix
$G_{n,M}\mathrel{\coloneqq}\E[(X-\widetilde m_{n,X})(X-\widetilde m_{n,X})^\top\mid\mathcal T_{n,k}]$
satisfies
\[
 \lambda_{\min}(G_{n,M})
 \ge\frac{\lambda_{\min}(B_n)^2}{\lambda_{\max}(\E[XX^\top])},
 \qquad \|G_{n,M}^{-1}\|=O_P(1).
\]
\end{lemma}

\begin{proof}
Fix $(C,E)=(c,e)$. By the range condition $\zeta_{n,k}\in\operatorname{range}(V_{n,k})$ in \Cref{ass:est}(b), the conditional Gaussian laws
$U_n\mid X=x$ have the same affine support for almost every $x$ in this conditional distribution. Indeed, on this conditional distribution,
\[
 \widehat C_n(X)=c+\E[\delta_n(X)\mid c,e,\mathcal T_{n,k}]+\zeta_n,
\]
so their means differ only by vectors in $\operatorname{range}(V_n)$.
Let $\pi_x(u)$ be their densities on that support, and let $\pi(u)\mathrel{\coloneqq}\E[\pi_X(u)\mid c,e,\mathcal T_{n,k}]$ be the mixture density.
Writing $P_{c,e}$ for the conditional law of $X$ given $(c,e,\mathcal T_{n,k})$, Bayes' rule gives the posterior law
\[
 P_{c,e}(dx\mid U_n=u)=\frac{\pi_x(u)}{\pi(u)}P_{c,e}(dx).
\]
For two independent draws from this posterior, the variance of $v^\top X$ is half the expected squared difference. Integrating this identity over $u$, whose density is $\pi$, cancels one factor of $\pi(u)$ in the denominator.
The conditional-variance identity gives, for any $v\in\R^d$,
\[
 \E[\Var(v^\top X\mid U_n,c,e,\mathcal T_{n,k})\mid c,e,\mathcal T_{n,k}]
 =\tfrac12\E\!\left[(v^\top\Delta X)^2
        \int\frac{\pi_X(u)\pi_{X'}(u)}{\pi(u)}\,du
         \,\middle|\,c,e,\mathcal T_{n,k}\right].
\]
Cauchy--Schwarz and the affinity of Gaussians with common covariance imply
\[
 \int\frac{\pi_x\pi_{x'}}{\pi}\,du
 \ge\left(\int\sqrt{\pi_x\pi_{x'}}\,du\right)^2
 =e^{-d_n^2/4}.
\]
Here Cauchy--Schwarz is applied to $\sqrt{\pi_x\pi_{x'}/\pi}$ and $\sqrt\pi$, using $\int\pi=1$. Completing the square on the common affine support gives $\int\sqrt{\pi_x\pi_{x'}}=e^{-d_n^2/8}$; squaring yields the displayed exponent.
By conditional total variance,
integration over $(C,E)$ gives $Q_n\succeq B_n$.
The intermediate bound is
\[
 Q_n\succeq\E[\Var(X\mid U_n,C,E,\mathcal T_{n,k})\mid\mathcal T_{n,k}]
 \succeq B_n.
\]

The zero-rank case is immediate because the range condition then gives $\zeta_n=0$: $U_n$ is a function of $(C,E)$ given the training data, and $d_n=0$, so $Q_n\succeq\Sigma_C=B_n$.

Write $s\mathrel{\coloneqq}\lambda_{\min}(\Sigma_C)>0$. The conditional centering of $\zeta_n$ gives
$\E[d_n^2\mid\mathcal T_{n,k}]=2\kappa_n$, while
$\frac12\E[\Delta X\Delta X^\top]=\Sigma_C$.
For the first identity, $\widehat C_n(X)-\widehat C_n(X')$ is the difference of two leakage terms that are independent and centered conditional on $(C,E,\mathcal T_{n,k})$. Thus the cross term in $d_n^2$ has expectation zero. The second identity follows from the same independent-pair representation of conditional variance.
Given any $\epsilon>0$, choose a fixed $A<\infty$ such that $\kappa_n\le A$ with probability at least $1-\epsilon$ for all sufficiently large $n$.
Since $\Delta X$ has a fixed distribution with finite second moment, choose $L<\infty$ with
$\E[\|\Delta X\|^2\1\{\|\Delta X\|^2>L\}]<s/2$.
On $\{\kappa_n\le A\}$, Markov's inequality gives
\[
 \begin{aligned}
 \E[\|\Delta X\|^2\1\{d_n^2>T\}\mid\mathcal T_{n,k}]
 &{}\le \E[\|\Delta X\|^2\1\{\|\Delta X\|^2>L\}]\\
 &\qquad{}+LP(d_n^2>T\mid\mathcal T_{n,k})\\
 &\le s/2+2LA/T\le s
 \end{aligned}
\]
for a sufficiently large fixed $T$, for example $T\ge4LA/s$.
On $\{d_n^2\le T\}$ the Gaussian weight is at least $e^{-T/4}$, while
\[
 \E[\Delta X\Delta X^\top\1\{d_n^2>T\}\mid\mathcal T_{n,k}]
 \preceq \E[\|\Delta X\|^2\1\{d_n^2>T\}\mid\mathcal T_{n,k}]I
 \preceq sI.
\]
Hence
$B_n\succeq e^{-T/4}(\Sigma_C-sI/2)\succeq e^{-T/4}sI/2$
on this event, proving the inverse bound: for every $\epsilon>0$, the smallest eigenvalue is bounded below by a fixed positive constant with probability at least $1-\epsilon$ for all sufficiently large $n$.

For exact prediction averaging define
$S_nX\mathrel{\coloneqq}\E_\xi[m_{n,X}(\widehat C_n(X)+\xi,E)\mid X,E,\mathcal T_{n,k}]$
and $G_{n,\infty}\mathrel{\coloneqq}\E[(X-S_nX)(X-S_nX)^\top\mid\mathcal T_{n,k}]$.
Conditional-expectation identities yield
$\E[X(X-S_nX)^\top\mid\mathcal T_{n,k}]=Q_n$.
To verify this, write $m\coloneqq m_{n,X}(U_n,E)$. Integrating first over the perturbation and then conditioning on $(U_n,E,\mathcal T_{n,k})$ gives
\[
 \E[X(S_nX)^\top\mid\mathcal T_{n,k}]
 =\E[Xm^\top\mid\mathcal T_{n,k}]
 =\E[mm^\top\mid\mathcal T_{n,k}].
\]
Subtracting from $\E[XX^\top]$ gives $Q_n$ by the definition of conditional variance.
For any $v$, Cauchy--Schwarz therefore gives
\[
 \|Q_nv\|^2\le\lambda_{\max}(\E[XX^\top])\,v^\top G_{n,\infty}v,
 \qquad
 \lambda_{\min}(G_{n,\infty})
 \ge\frac{\lambda_{\min}(Q_n)^2}{\lambda_{\max}(\E[XX^\top])}.
\]
For the first inequality, apply Cauchy--Schwarz to $(u^\top X)\,v^\top(X-S_nX)$ and maximize over unit vectors $u$. For the second, use $\|Q_nv\|\ge\lambda_{\min}(Q_n)\|v\|$ and minimize over unit vectors $v$.
Finally, $\E[\widetilde m_{n,X}\mid X,E,\mathcal T_{n,k}]=S_nX$, so
$G_{n,M}=G_{n,\infty}+\E[\Var(\widetilde m_{n,X}\mid X,E,\mathcal T_{n,k})\mid\mathcal T_{n,k}]\succeq G_{n,\infty}$.
The cross terms vanish because $\widetilde m_{n,X}-S_nX$ has conditional mean zero given $(X,E,\mathcal T_{n,k})$. The common perturbation marginals suffice for this identity. Combining this inequality with $Q_n\succeq B_n$ proves the stated bound for $G_{n,M}$.
\end{proof}
}
\section{Proofs}
\label{app:proofs}

\subsection[Proof of identification]{Proof of \Cref{thm:id}}
\label{app:proof-identification}
\begin{proof}
Define $W\coloneqq Y-\mu_E(X)$. Then $Y=\beta_0^\top X+q_E(X)+W$ and $\E[W\mid X,E]=0$.
Under \restrictionabbr, $q_E(X)$ is measurable with respect to $(C(X),E)$ using $q_e(X) = q_{e_0}(X) + C_e(X)$ for all $e$ with the convention $C_{e_0}(X)\mathrel{\coloneqq}0$, and
$\E[W\mid C(X),E]=0$ by the tower property. Therefore,
$\E[Y \mid C(X), E] = \beta_0^\top \E[X \mid C(X), E] + q_E(X)$. Subtracting this from $Y=\beta_0^\top X+q_E(X)+W$ yields
$\widetilde Y=\beta_0^\top\widetilde X+W$, and
it follows that $\E[\widetilde X\widetilde Y]=\Sigma_C\beta_0$, since $\E[\widetilde XW]=0$ by $(X,E)$-measurability of $\widetilde X$.
Positivity completes the proof.
\end{proof}

\subsection[Bias without the \restrictionname]{Bias without the \restrictionname}
\label{app:bias-without-ir}
\begingroup
Under the setting of \Cref{sec:problem_setting}, define
\[
r_E(X)\coloneqq q_E(X)-\E[q_E(X)\mid C(X),E].
\]
Subtracting the conditional expectation of $Y$ given $(C(X),E)$ from $Y=\beta_0^\top X+q_E(X)+W$ gives
\[
\widetilde Y=\beta_0^\top\widetilde X+r_E(X)+W,
\qquad
\E[\widetilde X\widetilde Y]=\Sigma_C\beta_0+\E[\widetilde Xr_E(X)],
\]
where we used $\E[W\mid C(X),E]=0$ and $\E[\widetilde XW]=0$ by the tower property. Under positivity, multiplying by $\Sigma_C^{-1}$ yields
\begin{equation}
\label{eq:bias-without-ir}
\beta_{\mathrm{\methodabbr}}-\beta_0=\Sigma_C^{-1}\E[\widetilde Xr_E(X)].
\end{equation}

For the sensitivity bound in \Cref{sec:sensitivity}, suppose $q_{e_0}(X)\in L^2$. Since $q_E(X)-q_{e_0}(X)$ is determined by $(C(X),E)$,
\[
r_E(X)=q_{e_0}(X)-\E[q_{e_0}(X)\mid C(X),E].
\]
The least-squares property of conditional expectation gives
\[
\E[r_E(X)^2]
\le \E\!\left[\bigl(q_{e_0}(X)-\E[q_{e_0}(X)\mid C(X)]\bigr)^2\right]
=\delta_{\mathrm{\restrictionabbr}}^2.
\]
For every unit vector $v\in\R^d$, \eqref{eq:bias-without-ir}, Cauchy--Schwarz, and the symmetry of $\Sigma_C^{-1}$ imply
\[
\bigl|v^\top(\beta_{\mathrm{\methodabbr}}-\beta_0)\bigr|
=
\left|\E\!\left[
(v^\top\Sigma_C^{-1}\widetilde X)r_E(X)
\right]\right|
\le \{v^\top\Sigma_C^{-1}v\}^{1/2}\{\E[r_E(X)^2]\}^{1/2}.
\]
Taking the supremum over unit vectors yields
\[
\|\beta_{\mathrm{\methodabbr}}-\beta_0\|_2
\le \frac{\{\E[r_E(X)^2]\}^{1/2}}{\sqrt{\lambda_{\min}(\Sigma_C)}}
\le \frac{\delta_{\mathrm{\restrictionabbr}}}{\sqrt{\lambda_{\min}(\Sigma_C)}}.
\]
\par
\endgroup

\textbf{Example: Equality without \restrictionabbr.}
Let $H,U_X,U_Y$ be mutually independent standard normal variables,
independent of $E$. For constants $a_e\in\mathbb R$, consider the SCM
\[
X\mathrel{\coloneqq}H+U_X,
\qquad
Y\mathrel{\coloneqq}\beta_0X+H^2-1+a_E+U_Y.
\]
Since $\E[H^2\mid X=x]=1/2+x^2/4$, we have
\[
q_e(x)=a_e+\frac{x^2-2}{4},
\qquad
C_e(x)=a_e-a_{e_0}.
\]
Thus $C(X)$ is constant, $\widetilde X=X$, and $\Sigma_C=2>0$.
The remaining nuisance is
\[
r_E(X)=\frac{X^2-2}{4},
\]
which is nonzero almost surely, so \restrictionabbr{} fails.
Nevertheless, symmetry of $X$ gives
\[
\E[\widetilde Xr_E(X)]
=\frac14\bigl(\E[X^3]-2\E[X]\bigr)=0.
\]
The bias identity~\eqref{eq:bias-without-ir} therefore yields
$\beta_{\mathrm{\methodabbr}}=\beta_0$ despite the violation of
\restrictionabbr{}: the remaining nuisance $r_E(X)$ is nonconstant but
uncorrelated with the residualized treatment $\widetilde X$.

\subsection[Proof of the unconfounded-environment result]{Proof of \Cref{prop:clean}}
\label{app:proof-clean}
\begin{proof}
If $e_\star=e_0$, then $q_{e_0}(X)$ is constant. Otherwise, $C_{e_\star}(X)=\kappa_\star-q_{e_0}(X)$, so $q_{e_0}(X)$ is determined by $C(X)$. Thus \restrictionabbr{} holds, and \Cref{thm:id} gives the result.
\end{proof}

\subsection[Proof of recoverable confounding]{Proof of \Cref{prop:recoverable}}
\label{app:proof-recoverable}
\begin{proof}
The outcome equation and $\E[U_Y\mid X,E]=0$ give $$q_E(X) =\E[\psi(H,E)\mid X,E] =\E[\psi(H,E)\mid Z_C,E]
=\rho_E(Z_C),$$ where we used $Z_C=F(X)$ almost surely
and $X\indep H\mid(Z_C,E)$. By environment overlap,
 $q_e(X)=\rho_e(Z_C)$ holds almost surely for every $e$, and hence $C(X)=\Delta(Z_C)$. Injectivity of $\Delta$ makes $Z_C$ measurable with respect to $C(X)$, proving \restrictionabbr, and also gives $\Sigma_C = \E[\Var(X\mid C(X),E)]=\E[\Var(X\mid Z_C,E)]=\Sigma_{Z_C} \succ 0$, since conditioning on $(C(X),E)$ is equivalent to conditioning on $(Z_C,E)$. Therefore, \Cref{ass:cs,ass:pos} hold, and \Cref{thm:id} completes the proof.
\end{proof}

\subsection[Proof of smoothed consistency]{Proof of \Cref{thm:consistency-full}}
\label{app:proof-consistency}
Cross-fitting generally makes the moment estimates from different folds dependent. We establish convergence separately within each fold and combine these limits using that the number of folds is fixed.
{\begin{proof}
Fix a fold, condition on $\mathcal T_{n,k}$, and suppress $k$. All expectations in the first three steps are over a fresh observation and the prediction perturbations, conditional on this training data.

\emph{Step 1: recovery of the confounding term.}
Under \restrictionabbr, $q_E(X)=\varphi_E(C(X))$ is square integrable. For every $\epsilon>0$, choose bounded Lipschitz functions $b_e$ such that
$\E[(q_E(X)-b_E(C(X)))^2]<\epsilon$; such functions are dense in $L^2$ of the finite-dimensional contrast distribution, separately in each of the finitely many environments. Let $L_b$ bound their Lipschitz constants. Because
$\E\|U_n-C(X)\|^2=\rho_n^2+\operatorname{tr}(V_n)$,
the projection property gives
\[
 \E[(q_E-m_{n,q}(U_n,E))^2]
 \le\E[(q_E-b_E(U_n))^2]
 \le 2\epsilon+2L_b^2\{\rho_n^2+\operatorname{tr}(V_n)\}.
\]
Part (a) of \Cref{ass:est}, followed by $\epsilon\downarrow0$, makes this $o_P(1)$.
For the average $\widetilde m_{n,q}\mathrel{\coloneqq}M^{-1}\sum_jm_{n,q}(\widehat C_n(X)+\xi_j,E)$, Jensen's inequality and the common Gaussian marginals give
\begin{equation}
\label{eq:smoothing-confounding-recovery}
 \E[(q_E-\widetilde m_{n,q})^2]
 \le\frac1M\sum_j\E[(q_E-m_{n,q}(\widehat C_n(X)+\xi_j,E))^2]=o_P(1).
\end{equation}

\emph{Step 2: the population residual equation.}
Linearity of conditional expectation and $\E[W\mid X,E]=0$ imply
$m_{n,Y}=\beta_0^\top m_{n,X}+m_{n,q}$.
Using the same prediction perturbations for $X$ and $Y$ preserves this identity after averaging. Thus the ideal residuals satisfy
\[
 R_X\mathrel{\coloneqq}X-\widetilde m_{n,X},\qquad
 R_Y\mathrel{\coloneqq}Y-\widetilde m_{n,Y}
     =\beta_0^\top R_X+(q_E-\widetilde m_{n,q})+W.
\]
Since the perturbations are independent of $W$ conditional on $(X,E)$,
$\E[R_XW]=0$.
By \Cref{lem:smoothing-overlap}, $G_{n,M}=\E[R_XR_X^\top]$ has inverse norm $O_P(1)$. Moreover, $\E\|R_X\|^2\le4\E\|X\|^2$, by the $L^2$ contraction property and Jensen's inequality. Therefore the score $\E[R_X(R_Y-\beta_0^\top R_X)]$ vanishes by \Cref{eq:smoothing-confounding-recovery} and Cauchy--Schwarz.
The population regression coefficient $\beta_{n,M}\mathrel{\coloneqq}G_{n,M}^{-1}\E[R_XR_Y]$ obeys the sharper bound
\[
 \|\beta_{n,M}-\beta_0\|
 \le\frac{\|q_E-\widetilde m_{n,q}\|_{L^2}}
              {\sqrt{\lambda_{\min}(G_{n,M})}}=o_P(1).
\]

\emph{Step 3: empirical ideal moments.}

We prove the foldwise weak laws by truncation, using only second moments of $X$ and $Y$. For $L>0$ and $Z\in\{X,Y\}$, set $Z^{(L)}\coloneqq Z\1\{\|Z\|\le L\}$ and define
\[
 m_{n,Z}^{(L)}(u,e)\coloneqq \E[Z^{(L)}\mid U_n=u,E=e,\mathcal T_{n,k}],\qquad
 R_Z^{(L)}\coloneqq Z^{(L)}-\frac1M\sum_{j=1}^M
 m_{n,Z}^{(L)}(\widehat C_n(X)+\xi_j,E).
\]
The conditional expectations have norm at most $L$, so $\|R_Z^{(L)}\|\le2L$. By $L^2$ contraction, Jensen's inequality, and the common perturbation marginals,
\[
 \begin{aligned}
 \E\|R_Z-R_Z^{(L)}\|^2
 &\le 2\E\|Z-Z^{(L)}\|^2
   +\frac2M\sum_{j=1}^M
     \E\bigl\|(m_{n,Z}-m_{n,Z}^{(L)})(\widehat C_n(X)+\xi_j,E)\bigr\|^2\\
 &\le4\E[\|Z\|^2\1\{\|Z\|>L\}].
 \end{aligned}
\]
The final bound is deterministic and uniform in $n$, $k$, and the training data, because the fresh observation is independent of the training data. Moreover, both $\E\|R_Z\|^2$ and $\E\|R_Z^{(L)}\|^2$ are at most $4\E\|Z\|^2$.

Let $A_i$ be any scalar entry of $R_{X,i}R_{X,i}^\top$ or $R_{X,i}(R_{Y,i}-\beta_0^\top R_{X,i})$, and let $A_i^{(L)}$ replace each residual by its truncated version. Define
\[
 \eta_L\coloneqq \sum_{Z\in\{X,Y\}}\E[\|Z\|^2\1\{\|Z\|>L\}]\longrightarrow0.
\]
Cauchy--Schwarz bounds the error of each product. For example,
\[
 \E\|R_XR_X^\top-R_X^{(L)}(R_X^{(L)})^\top\|
 \le \bigl(\E\|R_X-R_X^{(L)}\|^2\bigr)^{1/2}
 \left\{\bigl(\E\|R_X\|^2\bigr)^{1/2}
       +\bigl(\E\|R_X^{(L)}\|^2\bigr)^{1/2}\right\}.
\]
The same bound for cross-products gives, for a deterministic constant $c$ independent of $n$, $k$, $L$, and the training data,
\[
 |A_i^{(L)}|\le cL^2,\qquad
 \E[|A_i-A_i^{(L)}|\mid\mathcal T_{n,k}]\le c\sqrt{\eta_L}.
\]
Conditional on $\mathcal T_{n,k}$, the held-out rows are independent. Chebyshev's inequality for their bounded moments $A_i^{(L)}$ and Markov's inequality for the truncation errors therefore give, for each $\epsilon>0$,
\[
 P\!\left(\left|\frac1{|I_k|}\sum_{i\in I_k}
 \{A_i-\E[A_i\mid\mathcal T_{n,k}]\}\right|>\epsilon
 \,\middle|\,\mathcal T_{n,k}\right)
 \le c_\epsilon\left\{\frac{L^4}{|I_k|}+\sqrt{\eta_L}\right\},
\]
where $c_\epsilon$ is independent of $n$, $k$, $L$, and the training data. First let $|I_k|\to\infty$ with $L$ fixed, then let $L\to\infty$. The bound also holds after averaging over the training data, proving the foldwise weak laws.

For environment-stratified folds, apply the same argument under the fixed law conditional on each environment. The within-environment sample sizes diverge, and their proportions converge to $P(E=e)$. The conditional moment bounds make the error from replacing these proportions by their limits $o_P(1)$.
Pooling the fixed number of folds gives
the following limits, using the vanishing population score from Step 2 for the second one:
\[
 \frac1n\sum_i R_{X,i}R_{X,i}^\top
   -\sum_k\frac{|I_k|}{n}G_{n,k,M}=o_P(1),\qquad
 \frac1n\sum_i R_{X,i}(R_{Y,i}-\beta_0^\top R_{X,i})=o_P(1).
\]
By \Cref{lem:smoothing-overlap} and fixed $K$, the convex combination of Gram matrices has inverse norm $O_P(1)$, since
\[
 \begin{aligned}
 \lambda_{\min}\!\left(\sum_k\frac{|I_k|}{n}G_{n,k,M}\right)
 &\ge\min_{1\le k\le K}\lambda_{\min}(G_{n,k,M}),\\
 \left\|\left(\sum_k\frac{|I_k|}{n}G_{n,k,M}\right)^{-1}\right\|
 &\le\max_{1\le k\le K}\|G_{n,k,M}^{-1}\|=O_P(1).
 \end{aligned}
\]

\emph{Step 4: transfer to fitted regressions.}
Let $\nu_{Z,i}$ be the average of $f_{n,k,Z}-m_{n,k,Z}$ over the prediction draws for row $i\in I_k$. Part (c) of \Cref{ass:est} and Jensen's inequality give $\E[\|\nu_{Z,i}\|^2\mid\mathcal T_{n,k}]=o_P(1)$. Conditional Markov's inequality then implies $n^{-1}\sum_i\|\nu_{Z,i}\|^2=o_P(1)$ for $Z=X,Y$.
Since $\widehat X_i=R_{X,i}-\nu_{X,i}$ and $\widehat Y_i=R_{Y,i}-\nu_{Y,i}$, empirical Cauchy--Schwarz shows that replacing the ideal residuals by the fitted ones changes the Gram and score moments by $o_P(1)$. Consequently,
\[
 P(\widehat G_n\succ0)\to1,\quad
 \|\widehat G_n^{-1}\|=O_P(1),\quad
 \widehat s_n\coloneqq n^{-1}\sum_i\widehat X_i
             (\widehat Y_i-\beta_0^\top\widehat X_i)=o_P(1).
\]
Finally,
$\widehat\beta_{\mathrm{SOVR}}-\beta_0
=(\widehat G_n+\lambda_nI)^{-1}(\widehat s_n-\lambda_n\beta_0)=o_P(1)$.
\end{proof}}

\subsection[Proof of low-rank \MakeLowercase{\restrictionname}]{Proof of \Cref{thm:lowrank}}
\label{app:proof-lowrank}
We first prove a generalization of \Cref{thm:lowrank} that allows environment-specific intercepts and does not require a specified latent variable.
\begin{proposition}[General low-rank \MakeLowercase{\restrictionname}]
\label{prop:lowrank-general}
Under the setting of \Cref{sec:problem_setting}, let $g:\R^d\to\R^r$ be a measurable function. Suppose that for every $e\in\mathcal E$ there are an intercept $\alpha_e\in\R$ and loadings $\lambda_e\in\R^r$ such that
\[
q_e(X)=\alpha_e+\lambda_e^\top g(X)
\qquad P_X\text{-almost surely}.
\]
Fix a reference environment $e_0$, define $L_{e,\cdot}\coloneqq(\lambda_e-\lambda_{e_0})^\top$ for $e\ne e_0$, and set $\Sigma_g\coloneqq\E[\Var(X\mid g(X),E)]$.
If $\lambda_{e_0}\in\operatorname{rowspan}(L)$ and $\Sigma_g\succ0$, then \Cref{ass:cs,ass:pos} hold and $\beta_{\mathrm{\methodabbr}}=\beta_0$.
\end{proposition}
\begin{proof}
Let $\delta_\alpha
    \coloneqq 
    \bigl(\alpha_e-\alpha_{e_0}\bigr)_{e\ne e_0}$.
The contrast satisfies $C(X)=\delta_\alpha+Lg(X)$.
Since $\lambda_{e_0}\in\operatorname{rowspan}(L)$, there exists a $b\in\mathbb{R}^{|\mathcal E|-1}$
such that $\lambda_{e_0}=L^\top b$. It follows that
\[
    q_{e_0}(X)
    =
    \alpha_{e_0}+b^\top\bigl(C(X)-\delta_\alpha\bigr),
\]
which proves \restrictionabbr. Moreover, since $C(X)$ is a function of $g(X)$, the conditional variance
decomposition gives
\[
    \Sigma_C
    =\E[\Var(X\mid C(X),E)]
    \succeq
    \E[\Var(X\mid g(X),E)]
    =\Sigma_g\succ0.
\]
Thus positivity holds and \Cref{thm:id} gives $\beta_{\mathrm{\methodabbr}}=\beta_0$.
\end{proof}

\begin{proof}[Proof of \Cref{thm:lowrank}]
In \eqref{eq:vector-tracer-example}, define $g(x)\coloneqq\E[H\mid X=x]$ and set $\alpha_e\mathrel{\coloneqq}0$ for every $e$. The common conditional expectation $\E[H\mid X,E]=\E[H\mid X]$ and $\E[U_Y\mid X,E]=0$ imply $q_e(X)=\lambda_e^\top g(X)$. The two conditions in \Cref{thm:lowrank} are therefore those of \Cref{prop:lowrank-general}, which proves the claim.
\end{proof}

\section{Experimental Details}
\label{app:dgps}

\subsection{Synthetic SCMs}

In each synthetic setting from \Cref{sec:synthetic_experiments}, the environment label $E$ is sampled uniformly from a finite set of environments $\mathcal E$. All exogenous noises ($U_X$, $U_H$, $U_R$, $U_Y$) are mutually independent standard Gaussians of the appropriate dimension, independent of $E$. For the latent-loading experiment (a) and the recoverable-state experiment (c), structural functions and parameters are fixed across replications. For the MLP experiments ((b) and (d)--(f)), they are redrawn for each replication. In all settings, they remain fixed across sample sizes within a replication. Every MLP used to generate the synthetic data has hidden-layer width $32$ and tanh activations: for hidden layer $\ell$, $a_\ell\mathrel{\coloneqq}\tanh(a_{\ell-1}W_\ell+b_\ell)$, with entries of $W_\ell$ drawn from $N(0,1/d_{\ell-1})$ and entries of $b_\ell$ from $N(0,.1^2)$; the output layer is linear with analogous distributions for the weights and biases.

\textbf{(a) Latent loadings.}

This setting has two binary confounders, a treatment $X\in\R^4$, and four environments. To specify the model, independently draw $\beta_0\sim N(0,I_4)$ and independent vectors $v_1,v_2\sim N(0,I_2)$, and set $a_j\coloneqq v_j/\|v_j\|_2$.
  Taking $e_0=0$ as the reference, independently draw $v_0\sim N(0,I_2)$ and set
\[
\lambda_0\coloneqq2\frac{v_0}{\|v_0\|_2}.
\]
This gives a reference loading of norm $2$. We choose the other three loadings so that their differences from $\lambda_0$ span $\R^2$. The rows of $L\in\R^{3\times2}$ will specify these differences.
To construct $L$, independently draw matrices $G_L\in\R^{3\times2}$ and $G_R\in\R^{2\times2}$ with i.i.d.\ $N(0,1)$ entries and compute reduced QR decompositions $G_L=Q_LR_L$ and $G_R=Q_RR_R$. The columns of $Q_L$ and $Q_R$ are perpendicular unit vectors, so $Q_L^\top Q_L=Q_R^\top Q_R=I_2$. Set
\[
L\coloneqq Q_L
\begin{pmatrix}3&0\\0&2\end{pmatrix}Q_R^\top,
\qquad
\lambda_e\coloneqq\lambda_0+L_{e,\cdot}^{\top},\quad e\in\{1,2,3\},
\]
where $L_{e,\cdot}$ is the $e$th row of $L$. Thus $L_{e,\cdot}=(\lambda_e-\lambda_0)^\top$. The middle diagonal matrix fixes the singular values of $L$ at $3$ and $2$, while $Q_L$ and $Q_R$ determine the directions. Both singular values are positive, so $L$ has rank $2$. We draw $\beta_0$, $a_1,a_2$, and $\lambda_0,\ldots,\lambda_3$ once and hold them fixed across all replications and sample sizes.

For each observation, independently draw $H_1,H_2\sim\operatorname{Bernoulli}(1/2)$, $E\sim\operatorname{Unif}\{0,1,2,3\}$, $U_X^{(1)},U_X^{(2)}\sim N(0,I_2)$, and $U_Y\sim N(0,1)$. With $H\coloneqq(H_1,H_2)^\top$, generate
\[
\begin{aligned}
X^{(j)}&\coloneqq0.8(2H_j-1)a_j+\sigma_{H_j}U_X^{(j)},
\qquad j\in\{1,2\},\quad \sigma_0\coloneqq0.7,\quad\sigma_1\coloneqq1.3,\\
X&\coloneqq\bigl((X^{(1)})^\top,(X^{(2)})^\top\bigr)^\top,\\
Y&\coloneqq\beta_0^\top X+\lambda_E^\top H+0.5U_Y.
\end{aligned}
\]
Hence $E$ is independent of $(H,X)$ and changes only the confounders' coefficients in the outcome equation.

The contrasts satisfy
\[
C(X)=L\E[H\mid X]
=L\begin{pmatrix}
P(H_1=1\mid X^{(1)})\\
P(H_2=1\mid X^{(2)})
\end{pmatrix}.
\]
Since $L$ has rank $2$, the contrasts determine both posterior probabilities, and $\lambda_0\in\operatorname{rowspan}(L)$ as required by \Cref{thm:lowrank}. The unequal variances make each posterior log-odds quadratic in its treatment block, with circular level sets. Conditioning on the posterior probabilities therefore leaves variation in every treatment direction, giving positivity.

\textbf{(b) One unconfounded environment.}
Here $d=4$, $d_H=2$, and $E\sim\operatorname{Unif}\{0,\ldots,4\}$. Draw $m_e\mathrel{\coloneqq}.3z_e$ and $b_e\mathrel{\coloneqq}.5v_e$ with $z_e\sim N(0,I_2)$ and $v_e\sim N(0,I_4)$, then set $H\mathrel{\coloneqq}m_E+U_H$ and $X\mathrel{\coloneqq}\operatorname{std}\{f_X(H)\}+b_E+U_X$, where $f_X:\mathbb R^2\to\mathbb R^4$ is a two-hidden-layer MLP and $\operatorname{std}$ denotes coordinatewise centering and scaling on an independent calibration sample of size $4{,}000$. A second two-hidden-layer MLP maps $(\operatorname{onehot}(E),H)$ to a scalar $\psi(H,E)$. We subtract its empirical mean and divide by its empirical standard deviation, computed on the same calibration sample pooled across all five environments, to obtain $\widetilde\psi(H,E)$. We use $Y\mathrel{\coloneqq}\beta_0^\top X+3\widetilde\psi(H,E)\mathbf 1\{E\ne3\}+.5U_Y$, with $\beta_0\sim N(0,I_4)$. Thus environment $3$ is exactly unconfounded, while the Gaussian $U_X$ gives all environments common support.
Common support does not imply positivity (\Cref{ass:pos}), which is not guaranteed by this construction.

\textbf{(c) Recoverable confounding state.}
Let $E\sim\operatorname{Unif}\{0,1,2,3\}$, $R\mathrel{\coloneqq}\exp(.6U_R-.18)$, and $X\mathrel{\coloneqq}RV_E\in\mathbb R^4$ with $\|V_e\|=1$. Relative to the uniform distribution on the unit sphere, $V_e$ has density $1+.5\theta_e^\top v$, where $\theta_e\mathrel{\coloneqq}(\cos(2\pi e/4),\sin(2\pi e/4),0,0)$. $V_E$ is drawn independently of $(R,U_H)$. Set $b_e\mathrel{\coloneqq}3\cos(2\pi e/4)$, $(d_0,d_1,d_2,d_3)\mathrel{\coloneqq}(-.5,-1/6,1/6,.5)$, $H\mathrel{\coloneqq}R^2+b_E R+.5U_H$, and $Y\mathrel{\coloneqq}\beta_0^\top X+H+d_E+.5U_Y$, with $\beta_0\sim N(0,I_4)$.
This is the example following \Cref{prop:recoverable} with latent state $Z_C=R=\|X\|$, and $X\indep H\mid(R,E)$.
Then $q_e(x)=\|x\|^2+b_e\|x\|+d_e$, the contrast is injective in $R$, and the environment probabilities satisfy $P(E=e\mid X=x)=\{1+.5\theta_e^\top x/\|x\|\}/4$ for $x\ne0$. Thus $P(E=e\mid X)\ge1/8$ almost surely for every $e$, so the environment overlap condition~\eqref{eq:environment_overlap} holds. Conditioning on $(C(X),E)$ is equivalent to conditioning on $(R,E)$. Since $R>0$ and $V_E$ has positive density on the entire unit sphere, every nonzero linear combination of $X=RV_E$ retains positive conditional variance, so \Cref{ass:pos} holds.

\textbf{(d)--(f) Shared decoder: outcome, treatment, and latent shifts.}
Panels (d)--(f) use $d=4$, $d_H=1$, and five environments. Let $\widetilde A_e\in\R^{d_H}$ and $\widetilde B_e\in\R^d$ be independent standard Gaussian draws, centered across environments and divided by the standard deviation of all their entries. With independent three-hidden-layer MLPs $f_H$, $f_X$, and $\psi_e:\mathbb R\to\mathbb R$, we generate $H\mathrel{\coloneqq}\operatorname{std}\{f_H(U_H)\}+t\widetilde A_E$, $X\mathrel{\coloneqq}\operatorname{std}\{f_X(H,U_X)\}+s\widetilde B_E$, and $Y\mathrel{\coloneqq}\beta_0^\top X+3\operatorname{std}\{\psi_E(H)\}+.5U_Y$, where $\beta_0=2z/\|z\|_2$ for an independent draw $z\sim N(0,I_4)$ and standardization uses $10{,}000$ independent calibration draws. The shift pairs $(t,s)$ are $(0,0)$, $(0,.3)$, and $(.3,.3)$, yielding (d) Outcome shifts, (e) Outcome and treatment shifts, and (f) Outcome, treatment, and confounder shifts. Within each replication the three settings share structural maps, exogenous draws, and learner seeds. The first has $E\indep(H,X)$; the latter two successively add $E\to X$ and $E\to H$ while retaining common latent and treatment mechanisms.

\subsection{Estimators and synthetic results}
\label{app:estimators}

Both \methodabbr{} estimators use an environment-balanced $50$--$50$ split between $\mathcal I_C$ and $\mathcal I_\beta$ and three-fold cross-fitting on $\mathcal I_\beta$. The first- and second-stage networks have three and two hidden layers of width $64$ and train for at most $200$ and $300$ epochs, respectively. D\methodabbr{} jointly fits $\widehat\mu_e=\widehat f+\widehat c_e$ on the full contrast split, using the shared features and separate output layers described in \Cref{app:estimation}. Training uses standardized inputs and outcomes and squared prediction loss, with an $\ell_2$ penalty of $\frac{10^{-4}}2\sum_{e\ne e_0}(\|w_e\|_2^2+b_e^2)$ on the contrast-output weights and biases. The shared features and reference-output layer are unpenalized. The second stage receives the vector $\widehat C(X)=(\widehat c_e(X))_{e\ne e_0}$. S\methodabbr{} divides the contrast split again, fits two members with penalty $10^{-2}$ and $15\%$ validation sets, averages their contrasts, and uses their disagreement covariance for Gaussian smoothing with $\tau=1$ and $M=8$ perturbations arranged in four antithetic pairs. Both stages use Adam: the first uses learning rate $5\times10^{-3}$ and batches of $256$; the second uses full batches, learning rate $10^{-2}$, and weight decay $10^{-4}$. Each second-stage fold fits a single network with $d+1$ outputs to predict the coordinates of $X$ and $Y$ jointly. Early stopping for the smoothed contrast learners uses patience $15$.

OLS ignores $E$; it linearly regresses $Y$ on $X$, including an intercept. FE subtracts environment means from $X$ and $Y$ before regression. Anchor regression uses environment indicators $\operatorname{onehot}(E)$ as anchors and $\gamma=2$. In designs with $E\indep X$, $\E[X\mid E]=\E[X]$ so centering within environments and applying the anchor transformation both leave centered $X$ unchanged at the population level. Environment-specific changes to $Y$ then have zero covariance with centered $X$. Thus OLS, FE, and anchor regression have the same population coefficient, explaining their similar finite-sample results. All plotted points are mean $\ell_2$ estimation error; shaded bands in \Cref{fig:synthetic-core} show $\pm$ one Monte Carlo standard error.

\textbf{Linear-plus-neural-network baseline (LNN).}\label{app:lnn} LNN jointly optimizes $(\beta,\theta)$ to minimize squared prediction error for $\beta^\top X+f_\theta(X,E)$. The neural network $f_\theta$ receives standardized $X$ and environment indicators and has two width-$64$ ReLU layers. The linear coefficient is unpenalized. Validation outcome error selects network weight decay from $\{0,10^{-4},10^{-2}\}$, initialization of $\beta$ from FE or zero, and the stopping epoch, with patience $30$. Adam uses learning rate $.001$ and batches of $1{,}024$; training lasts at most $200$ epochs on synthetic data and $120$ on real data.

Network weight decay, coefficient initialization, and stopping epoch are selected using a validation split within $80\%$ of the observations. The selected model is then refitted on the full sample for coefficient evaluation. Splits keep each user in one subset when user identifiers are available. Standardization is fitted on the corresponding training sample. Since $f_\theta(X,E)$ can represent linear dependence on $X$, the prediction objective alone does not identify $\beta$; LNN reports the coefficient selected by this training procedure.

\subsection[Supplementary synthetic comparisons]{Supplementary synthetic comparisons}
\label{app:synthetic-comparisons}
\begin{figure}[t]
\centering
\resizebox{\linewidth}{!}{\input{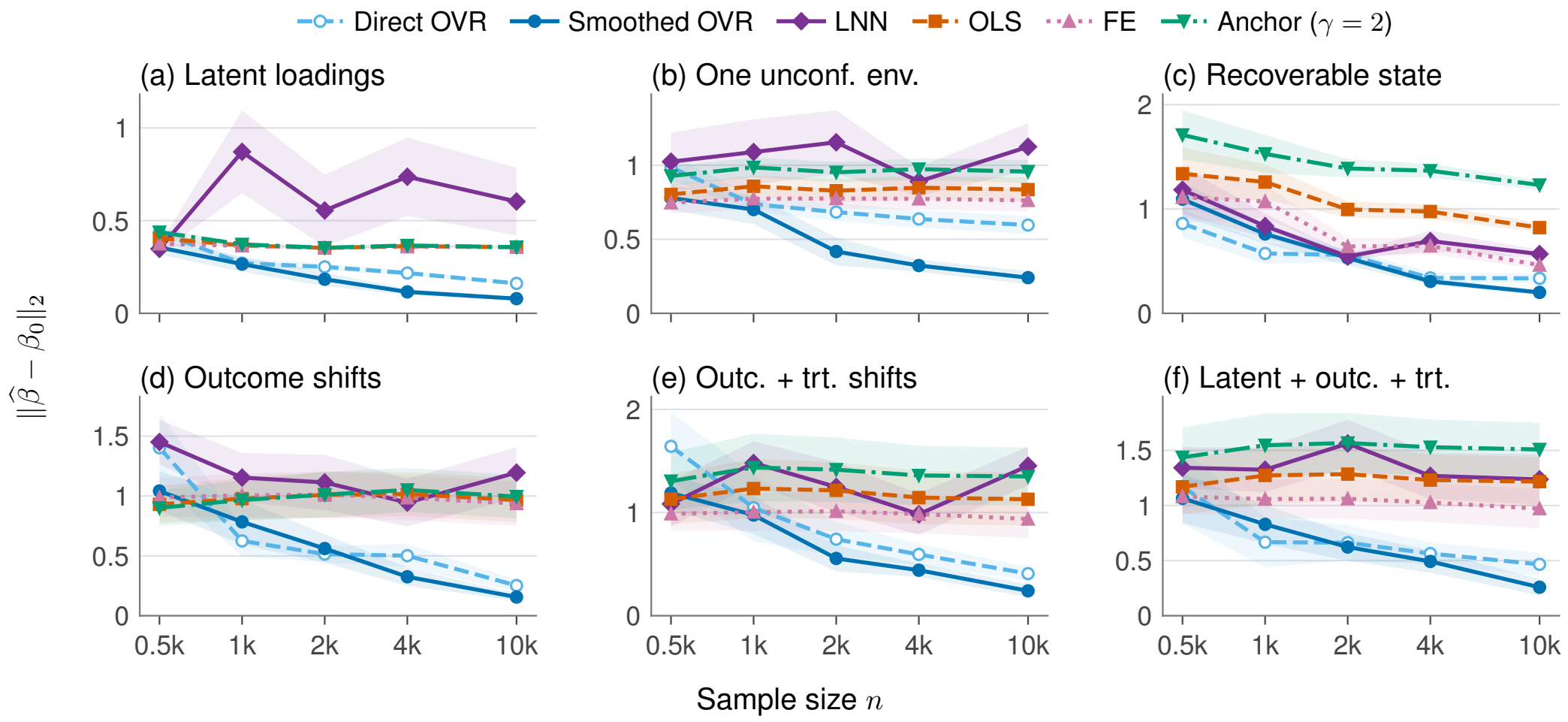}}
\caption{Synthetic comparison at $n\in\{0.5,1,2,4,10\}\mathrm{k}$, including Direct \methodabbr{}. Means and $\pm$ one Monte Carlo s.e.\ over ten paired replications.}
\label{fig:synthetic-large}
\end{figure}

\Cref{fig:synthetic-large} adds D\methodabbr{} to the main
comparison at $n\in\{500,1000,2000,4000,10000\}$, using
the same observations and ten replications as
\Cref{fig:synthetic-core}.

\subsection{Controlled component ablation}
\label{app:ablation}
We compare five nested variants at $n=10{,}000$ in all six synthetic settings, with ten paired replications per setting: D\methodabbr{} with one contrast fit, the same fit with stronger contrast-head regularization, the regularized fit with a $15\%$ validation holdout and early stopping, two such fits on disjoint halves of the contrast split averaged without smoothing ($\tau=0$), and the same ensemble with uncertainty smoothing ($\tau=1$). All variants share observations, the outer split, second-stage folds, and nuisance-learner seeds. The last two variants also share the fitted ensemble and disagreement estimate, so their comparison isolates smoothing. For pairing, the ablation D\methodabbr{} uses S\methodabbr's second-stage folds and can therefore differ from D\methodabbr{} run separately. The S\methodabbr{} estimates match those in \Cref{fig:synthetic-core} at $n=10{,}000$.

Let $n_\beta\mathrel{\coloneqq}|\mathcal I_\beta|$ be the number of observations used to estimate the coefficient, and define the residual second moment and cross-moment
\[
\widehat G\mathrel{\coloneqq}\frac{1}{n_\beta}\sum_{i\in\mathcal I_\beta}\widehat X_i\widehat X_i^\top,
\qquad
\widehat h\mathrel{\coloneqq}\frac{1}{n_\beta}\sum_{i\in\mathcal I_\beta}\widehat X_i\widehat Y_i.
\]
The implementation uses $\widehat G_\lambda\mathrel{\coloneqq}\widehat G+\lambda I$ with a numerical ridge $\lambda=10^{-10}$, so
\[
\widehat\beta-\beta_0=\widehat G_\lambda^{-1}\widehat r,
\qquad
\widehat r\coloneqq\widehat h-\widehat G_\lambda\beta_0.
\]
For $\widehat r\ne0$,
\[
\underbrace{\|\widehat\beta-\beta_0\|_2}_{\text{estimation error}}
=
\underbrace{\|\widehat r\|_2}_{\text{moment violation}}
\underbrace{\frac{\|\widehat G_\lambda^{-1}\widehat r\|_2}{\|\widehat r\|_2}}_{\text{amplification}}.
\]

Thus $\|\widehat r\|_2$ measures the violation of the empirical estimating equation at $\beta_0$, while $\|\widehat G_\lambda^{-1}\widehat r\|_2/\|\widehat r\|_2$ measures how much the inverse residual second-moment matrix scales this violation. The latter factor concerns the observed direction $\widehat r$, rather than the operator norm of $\widehat G_\lambda^{-1}$.

\begin{table}[H]
\centering
\caption{Component ablation at $n=10{,}000$. For each displayed quantity, entries are geometric means of the added-to-previous variant ratios over $60$ paired comparisons (six settings, ten replications each); values below one indicate reductions. ``Lower error'' counts pairs in which the added component reduces $\|\widehat\beta-\beta_0\|_2$.}
\label{tab:component-ablation}
\small
\setlength{\tabcolsep}{4pt}
\begin{tabular}{lcccc}
\toprule
Added component & $\|\widehat\beta-\beta_0\|_2$ & \shortstack{Lower\\error} & $\|\widehat r\|_2$ & $\dfrac{\|\widehat G_\lambda^{-1}\widehat r\|_2}{\|\widehat r\|_2}$ \\
\midrule
Contrast-head regularization & $0.926$ & $42/60$ & $0.774$ & $1.196$ \\
Validation and early stopping & $0.656$ & $42/60$ & $0.611$ & $1.074$ \\
Disjoint split ensemble ($\tau=0$) & $1.080$ & $25/60$ & $0.881$ & $1.227$ \\
Uncertainty smoothing ($\tau=1$) & $0.792$ & $44/60$ & $1.147$ & $0.691$ \\
\bottomrule
\end{tabular}
\end{table}

Validation and early stopping give the largest aggregate reductions in estimation error and $\|\widehat r\|_2$. The disjoint ensemble alone increases geometric-mean estimation error by $8.0\%$, while providing the disagreement estimate used for smoothing. Relative to this identical unsmoothed ensemble, smoothing reduces geometric-mean estimation error by $20.8\%$: $\|\widehat r\|_2$ increases by $14.7\%$, but $\|\widehat G_\lambda^{-1}\widehat r\|_2/\|\widehat r\|_2$ decreases by $30.9\%$.

Smoothing increases $\lambda_{\min}(\widehat G)$ in $58/60$ pairs and the positivity diagnostic $\widehat\gamma$ (\Cref{sec:diagnostics}) in $59/60$. In each of the six settings, smoothing yields lower arithmetic-mean estimation error and a paired geometric-mean error ratio below one.

\subsection[Sensitivity to the smoothing scale]{Sensitivity to the smoothing scale}
\label{app:tau-sweep}
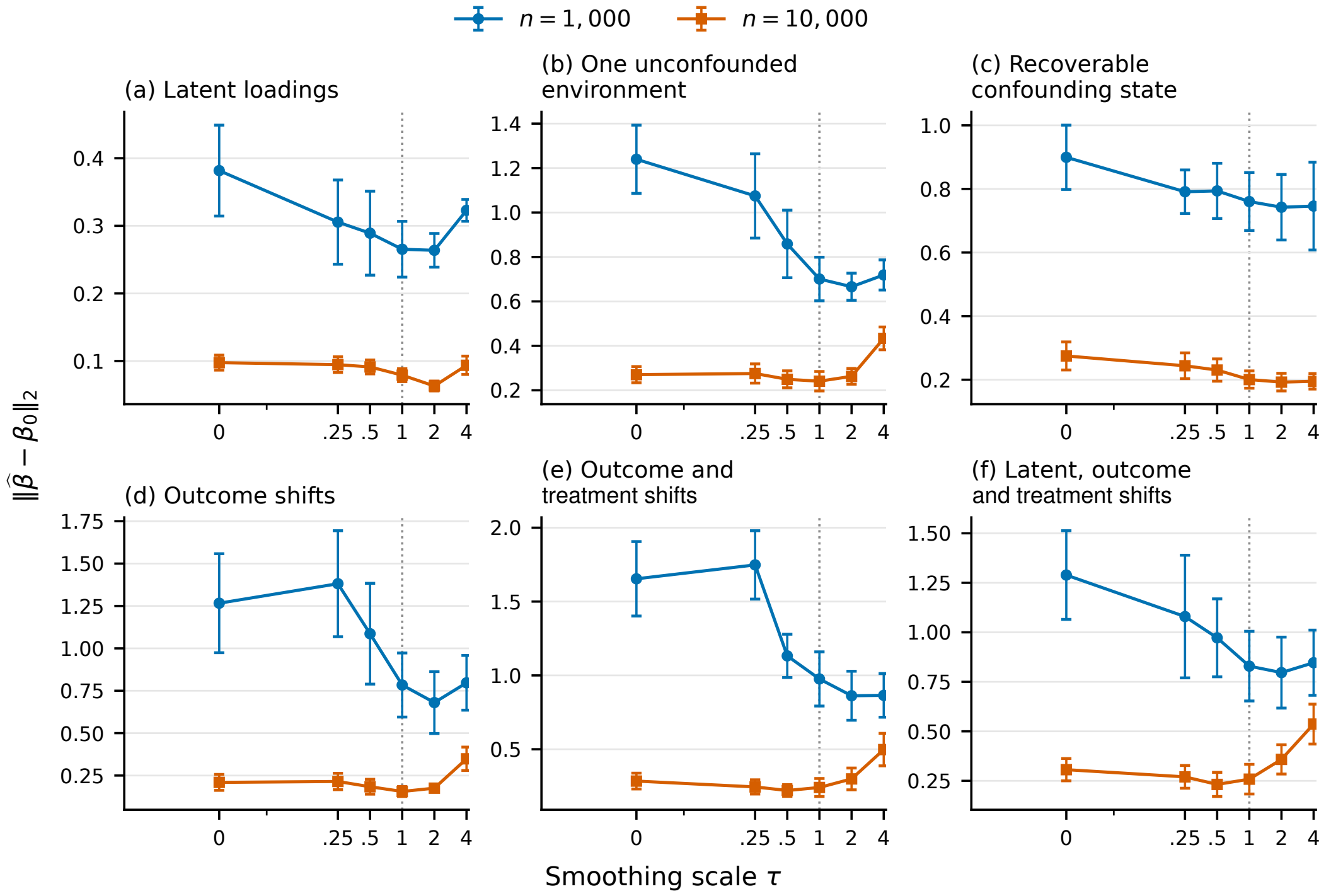
\begin{figure}[t]
\centering
\begin{tikzpicture}
\node[anchor=south west,inner sep=0] (plot) at (0,0) {\includegraphics[width=\linewidth]{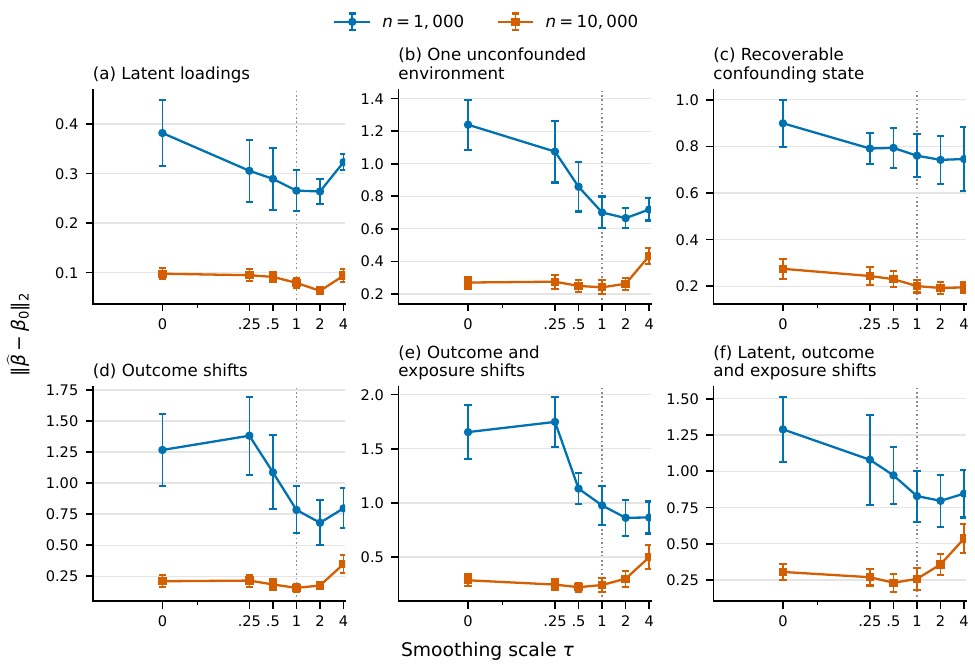}};
\begin{scope}[x={(plot.south east)},y={(plot.north west)}]
\fill[white] (0.40764231,0.42819292) rectangle (0.53959717,0.46040265);
\node[anchor=south west,inner sep=0,font=\sffamily\fontsize{0.01709402\linewidth}{0.01923077\linewidth}\selectfont] at (0.40871069,0.43505933) {treatment shifts};
\fill[white] (0.73061428,0.42819292) rectangle (0.90017598,0.46040265);
\node[anchor=south west,inner sep=0,font=\sffamily\fontsize{0.01709402\linewidth}{0.01923077\linewidth}\selectfont] at (0.73168266,0.43505933) {and treatment shifts};
\end{scope}
\end{tikzpicture}
\caption{Sensitivity of Smoothed \methodabbr{} to the smoothing scale. Points and bars show mean $\ell_2$ estimation error and $\pm$ one Monte Carlo standard error over ten paired replications. All scales reuse the same fitted contrast ensemble, second-stage folds, and second-stage random seeds. The dotted vertical line marks the default $\tau=1$; panels use separate linear error scales.}
\label{fig:tau-sweep}
\end{figure}

\begingroup
We vary the smoothing scale over $\tau\in\{0,0.25,0.5,1,2,4\}$ in all six synthetic settings at $n\in\{1{,}000,10{,}000\}$, with ten paired replications per setting and sample size. Each replication fits the two contrast learners once and reuses their average, disagreement estimate, second-stage folds, and second-stage random seeds at every scale. All other settings match S\methodabbr{} in \Cref{app:estimators}. The $\tau=0$ endpoint uses the same two-model ensemble without smoothing.

\Cref{fig:tau-sweep} summarizes the $720$ fits. At $n=1{,}000$, $\tau=2$ has the lowest observed mean error in all six settings, although its paired differences from $\tau=1$ are small relative to their Monte Carlo uncertainty. At $n=10{,}000$, the lowest means occur at $\tau=2$ in (a),(c), $\tau=1$ in (b),(d), and $\tau=0.5$ in (e),(f). The default $\tau=1$ has lower mean error than $\tau=0$ in all twelve setting--sample-size combinations. Larger perturbations can increase error: in (f) at $n=10{,}000$, mean error rises from $0.259$ at $\tau=1$ to $0.537$ at $\tau=4$. The sweep does not identify a single best scale across all settings and sample sizes.
\par
\endgroup

\subsection{KuaiRand-Pure held-out randomized reference}
\label{app:kuairand}
We restrict both assignment policies to the same application scenario, identified by the dataset field $\texttt{tab}=1$. The field identifies where an interaction occurred in the application; it is distinct from whether the video was assigned by the standard recommender or the random policy.
The cohort, training/reference splits and reference regression are described in \Cref{sec:real-data}. We retain only videos of type \texttt{NORMAL} with valid duration, dimensions, and upload dates, cap tag count at $6$, and require users to have an earlier interaction with a valid video on the same interface tab (\texttt{tab=1}). Both \methodabbr{} estimators use a user-disjoint $50$--$50$ split between $\mathcal I_C$ and $\mathcal I_\beta$, width-$64$ networks with three first-stage and two second-stage hidden layers trained for $120$ fixed first-stage and $160$ second-stage epochs, three second-stage folds, and batch size $512$ in the first stage; second-stage training uses full batches. The joint second-stage network, optimizer, learning rates, and second-stage weight decay are as in \Cref{app:estimators}. D\methodabbr{} fits one contrast network on the full contrast split with penalty $10^{-4}$; S\methodabbr{} fits two with penalty $10^{-2}$ and uses $\tau=1$ with $M=8$ perturbations arranged in four antithetic pairs.
\Cref{tab:kuairand-details} reports distances to the held-out randomized reference over the ten training/reference splits. Both \methodabbr{} variants have smaller distances than OLS, FE, and anchor regression in all ten paired comparisons. LNN uses the same training observations and scoring references, with user-disjoint validation.

\begin{table}[H]
\centering
\caption{KuaiRand-Pure: mean $\ell_2$ distance to the held-out randomized reference, with Monte Carlo standard errors in parentheses, over ten splits. The last two columns summarize the positivity diagnostic from \Cref{sec:diagnostics} across splits.}
\label{tab:kuairand-details}
\begin{tabular}{lccc}
\toprule
Method & Reference distance & Mean $\widehat\gamma$ & Minimum $\widehat\gamma$ \\
\midrule
D\methodabbr{} & $.0202\;(.0022)$ & $.339$ & $.287$ \\
S\methodabbr{} & $.0188\;(.0013)$ & $.397$ & $.351$ \\
OLS & $.1289\;(.0010)$ & --- & --- \\
FE & $.1039\;(.0009)$ & --- & --- \\
Anchor ($\gamma=2$) & $.1708\;(.0010)$ & --- & --- \\
LNN & $.0954\;(.0111)$ & --- & --- \\
\bottomrule
\end{tabular}
\end{table}

For D\methodabbr{}, the mean and minimum diagnostic values are $.339$ and $.287$; for S\methodabbr{}, they are $.397$ and $.351$. Even in the split with the smallest diagnostic, the fitted residual second moment is at least $28.7\%$ (D\methodabbr{}) or $35.1\%$ (S\methodabbr{}) of the within-environment variation in every linear treatment direction. These values do not suggest near-singularity of the fitted residual covariance relative to $\widehat\Sigma_E$, but they do not establish population positivity or have a calibrated pass/fail threshold.

\end{document}